\documentclass[10pt,twocolumn,letterpaper]{article}

\usepackage[pagenumbers]{config/iccv} 

\newif\ifshowcomments
\showcommentsfalse

\newcommand{\vale}[1]{\ifshowcomments\textbf{\textcolor{Mulberry}{[V: #1]}}\fi}

\usepackage{booktabs, multirow} 
\usepackage{soul}
\usepackage{xcolor,colortbl} 
\usepackage{changepage,threeparttable} 

\usepackage{multirow}
\usepackage{booktabs}

\definecolor{iccvblue}{rgb}{0.21,0.49,0.74}
\usepackage[pagebackref,breaklinks,colorlinks,allcolors=iccvblue]{hyperref}

\def\paperID{14} 
\def\confName{ICCV-CDEL}
\def\confYear{2025}

\title{Autoguided Online Data Curation for Diffusion Model Training}

\author{Valeria Pais\\
University of Glasgow\\
11 Chapel Ln, Glasgow G11 6EW, UK\\
{\tt\small v.pais-malacalza.1@research.gla.ac.uk}
\and
Luis Oala\\
Dotphoton\\
Nordstrasse 3, 6300 Zug, Switzerland\\
{\tt\small luis.oala@dotphoton.com}
\and
Daniele Faccio\\
University of Glasgow\\
11 Chapel Ln, Glasgow G11 6EW, UK\\
{\tt\small daniele.faccio@glasgow.ac.uk}
\and
Marco Aversa\\
Dotphoton\\
Nordstrasse 3, 6300 Zug, Switzerland\\
{\tt\small marco.aversa@outlook.com}
}

\begin{document}
\maketitle
\begin{abstract} 
The costs of generative model compute rekindled promises and hopes for efficient data curation. In this work, we investigate whether recently developed autoguidance and online data selection methods can improve the time and sample efficiency of training generative diffusion models. We integrate joint example selection (JEST) and autoguidance into a unified code base for fast ablation and benchmarking. We evaluate combinations of data curation on a controlled $2\text{-D}$ synthetic data generation task as well as $(3\times64^2)\text{-D}$ image generation. Our comparisons are made at equal wall-clock time and equal number of samples, explicitly accounting for the overhead of selection. Across experiments, autoguidance consistently improves sample quality and diversity. Early AJEST —applying selection only at the beginning of training— can match or modestly exceed autoguidance alone in data efficiency on both tasks. However, its time overhead and added complexity make autoguidance or uniform random data selection preferable in most situations. These findings suggest that while targeted online selection can yield efficiency gains in early training, robust sample quality improvements are primarily driven by autoguidance. We discuss limitations and scope, and outline when data selection may be beneficial.
\vspace{-.4cm}
\end{abstract}
\section{Introduction}
\label{sec:intro}
\begin{figure*}[h!]
    \centerline{\includegraphics[width=0.97\textwidth]{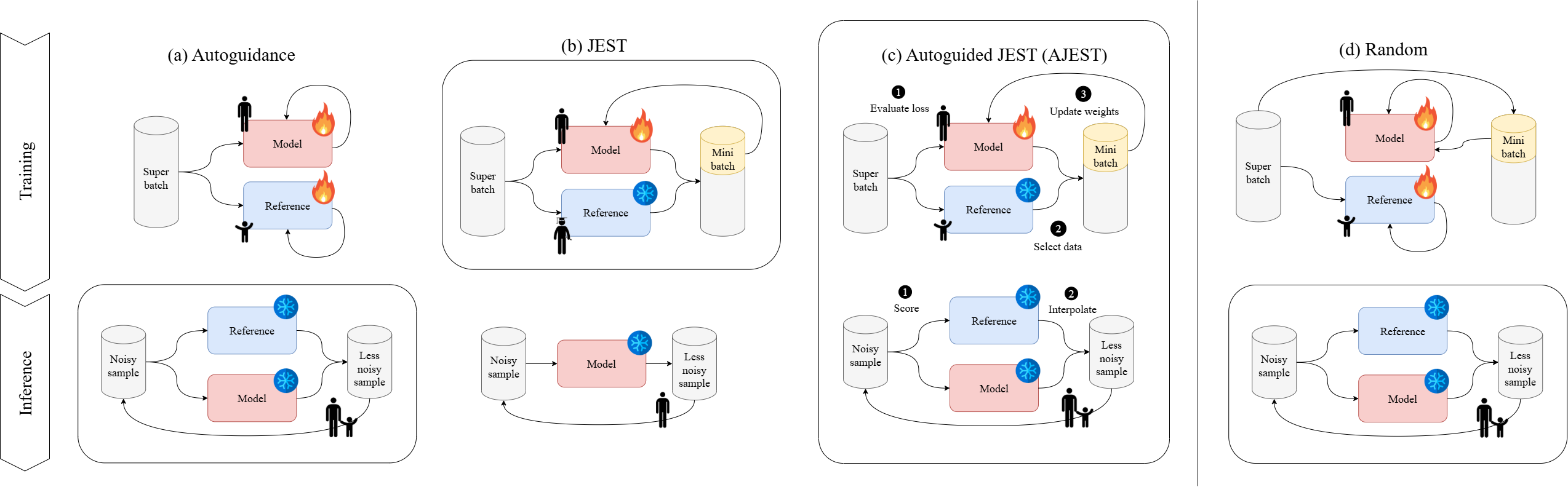}}
    \vspace{-3mm}\caption{Visual summary. (a) JEST: a large pre-trained foundational reference model is used to select data to train the main model, which can be smaller in size; inference is done using only the main model. (b) Autoguidance: a smaller reference model is trained for less iterations; both reference and main model score samples during the denoising process. (c) Autoguided JEST (AJEST): combination of JEST's training strategy and autoguidance's small less-trained reference and collaborative denoising process. (d) Random: JEST is replaced with uniform random data selection to act as a benchmarking baseline with independent training for the main model.}
    \label{fig:method}
    \vspace{-.3cm}
  \end{figure*}
Scaling trends have made modern generative modeling increasingly expensive, shifting attention from model- and hardware-centric efficiency towards data efficiency (e.g., \cite{2020KaplanScalingLanguage,2023ChertiScalingCLIP,2022SorscherBeatingScalingCNNs,oala2024dmlr,BISCHL2025101317}). Evidence suggests that targeted curation can improve performance by removing redundancy and noise \cite{2024GoyalAgnosticScaling}. Diffusion models, however, aim to approximate full data distributions, raising the question of whether data selection can help without harming diversity; recent work indicates pruning can help in some regimes \cite{2024BriqPruningDiffusion}.

We study online data selection for diffusion models through Autoguided JEST (AJEST), which integrates joint example selection (JEST) with autoguidance. We evaluate on a controlled 2-D task and Tiny ImageNet at $3\times64\times64$. Comparisons are made at equal wall-clock time and equal number of samples, explicitly accounting for selection overhead. 
Our focus is practicality: does targeted early selection provide efficiency gains, and how does it interact with autoguidance? 
\textbf{Our contributions and findings} are as follows:

\begin{enumerate}
    \item A unified code base to run combinations of autoguidance and JEST for online data curation of diffusion models\footnote{\url{https://github.com/0xInfty/Autoguided}}
    \item An evaluation harness for time and data efficiency of data curation during diffusion model training.
    \item Early AJEST can match or modestly exceed autoguidance in data efficiency on both a 2-D and $(3\times64^2)\text{-D}$; however, its time overhead and added complexity makes it less attractive for most applications.
    \item A transparent discussion of limitations and where data curation hits limits.
\end{enumerate}

\textbf{Related work.}\label{sec:related} Data selection and active learning span diversity-based approaches (e.g., k-center and submodular selections \cite{2018SenerSavarese,2020AgarwalContextual}) and gradient/importance-based methods \cite{2020MirzasoleimanCRAIG,2020AshAgarwalBADGE,2022PooladzandiAdaCore,2023ShinLCMat,2023XiaModerate}, as well as instance difficulty and loss-based strategies \cite{2018TonevaForgetting,2020FeldmanMemorization,2020ColemanSVP,2021PaulEL2NGradN,2018HanCoTeaching,2018JiangMentorNet,2023TanMoSo}. However, existing literature also suggests that random data selection can outperform these approaches in many settings \cite{2024OkanovicRS2}.

JEST \cite{2024EvansJEST} formalizes online joint selection using a learner–reference pair and has been shown capable of replacing knowledge distillation in contrastive setups \cite{2024UdandaraoACID}. For diffusion models, classifier-free guidance \cite{2022HoCFG} improves fidelity at the cost of diversity, whereas autoguidance \cite{2024KarrasAutoGuidance} leverages a weaker guide to boost quality while better preserving coverage. Our work examines how JEST-style online selection interacts with autoguidance for diffusion models under equal-time and equal-backprop comparisons.

\section{Methods}
\label{sec:setup}

We briefly review the core ingredients underlying the experiments, but we include details in \Cref{sec:implementation}.

\textbf{Diffusion models.} Diffusion models learn a family of scores over noise levels and iteratively denoise samples from Gaussian noise towards the data distribution \cite{2022KarrasEDM}. Sampling follows a probability‑flow ODE discretized over a noise (time) schedule; the model provides score evaluations that guide each denoising step. We adopt EDM/EDM2 conventions for preconditioning and training \cite{2024KarrasEDM2}.

Given the cost and scale of diffusion training, online data curation can appear appealing to reduce redundancy while preserving performance. Here we evaluate three regimes: an online learner–reference strategy (JEST) that adapts to training dynamics in combination with autoguidance (AJEST), a standalone autoguidance baseline, and a cheap random subset selection baseline (see \cref{fig:method}).

\textbf{JEST.} Joint Example Selection (JEST) \cite{2024EvansJEST} pairs a learner with a reference model to score examples in a super‑batch and draws a mini‑batch via softmax sampling over learnability scores. With super‑batch size $B$ and filtering ratio $f$, the update mini‑batch has size $b=(1-f)B$. The canonical learnability contrasts learner and reference losses and prioritizes examples that are easy to learn for the reference but not for the learner. JEST can act as implicit knowledge distillation in contrastive setups \cite{2024UdandaraoACID}. Exact scoring, chunked sampling logits, and our stability normalization are given in \Cref{sec:supp-ajest}.

\textbf{Autoguidance.} Classifier‑free guidance (CFG) \cite{2022HoCFG} improves fidelity at a diversity cost by mixing conditional/unconditional scores. Autoguidance \cite{2024KarrasAutoGuidance} instead leverages a weaker guide model (smaller or less‑trained) to produce a corrective signal that boosts quality while better preserving coverage; we follow their formulation and report both unguided and autoguided sampling. Exact guiding signal and collaborative sampling details are in \Cref{sec:supp-ajest}.

\textbf{AJEST.} We integrate JEST with autoguidance for diffusion models. A smaller guide (weaker version of the learner) serves as the JEST reference. Because the learner quickly surpasses the guide, we apply selection primarily at the beginning of training (Early AJEST) and then continue without selection; we also report a full‑selection variant for completeness. At inference, we evaluate both unguided sampling and autoguided collaborative sampling. 

\textbf{Random data selection.} As a cheap control, we uniformly sample a mini‑batch of size $b$ from each super‑batch of size $B$, matching the filtering ratio and schedule used by selective methods. This removes learner–reference scoring entirely, yields negligible compute and memory overhead, and provides a strong time‑efficiency baseline. The training loop and optimization settings are identical to those of AJEST and autoguidance; only the selection rule differs. 
In contrast with lowering the batch size, random data selection ensures some examples in the training dataset might never be seen by the learner model.

\section{Experiments}
\label{sec:experiments}
We compare methods under two budgets: (i) equal time budget, explicitly including selection overhead; and (ii) equal data budget. This allows us to evaluate both time- and data-efficiency.

\begin{figure}[t!]
  \centerline{\includegraphics[width=0.4\textwidth]{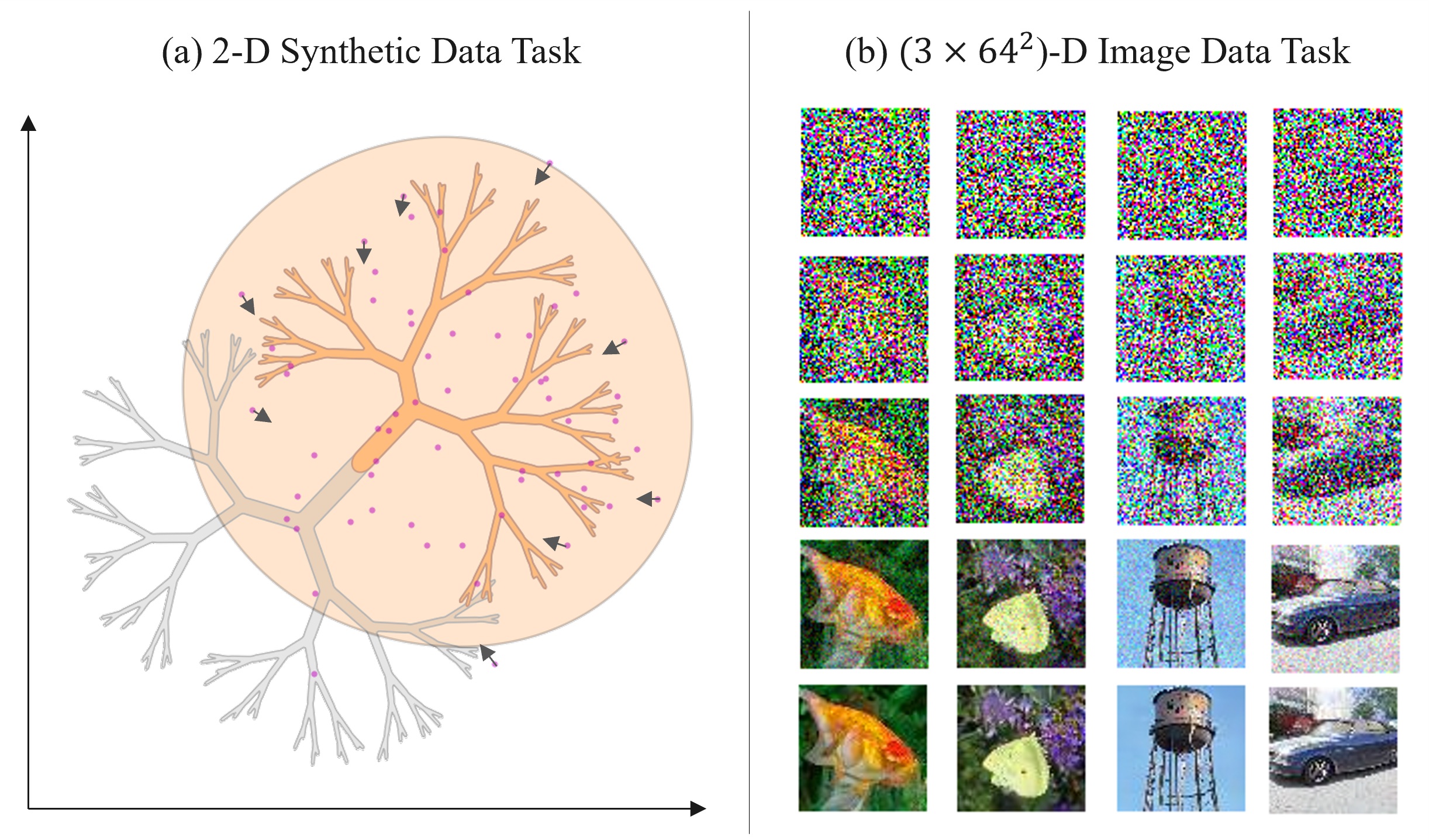}}
  \caption{Illustration of the data generation tasks. (a) Normal 2D $(x,y)$ samples (purple points) with noise level $\sigma$ (orange circle) are pushed towards the ground truth tree to obtain the final samples (gray points). (b) High-dimensional Normal samples are denoised to obtain natural images matching Tiny ImageNet's distribution.}
  \label{fig:tree}
  \vspace{-.6cm}
\end{figure}

\textbf{2-D Synthetic Data.} We first tested our methods on Karras et al.'s 2D tree task \cite{2024KarrasAutoGuidance}. The goal is to generate $(x,y)$ points with most of its probability density lying within a tree-shaped manifold (Figure~\ref{fig:tree}). This mimics low local dimensionality and hierarchical detail emergence in natural images \cite{2024KarrasAutoGuidance}. We train simple diffusion models to sample from one of two classes (upper half of the tree). 

We use Karras et al.'s toy model architecture \cite{2024KarrasAutoGuidance}. We trained a small reference model for 512 iterations and a larger main learner model for 4096 iterations, optionally applying data selection with mini-batch size $b=812$, super-batch size $B=8192$ and filtering ratio $f=0.8$. The loss function is evaluated on the whole batch of 8192 data points, but backpropagation is only executed for 20\% of those points. Details on the model and distribution are provided in \Cref{sec:supp-2d}.
Evaluation uses loss- and coverage-based metrics (MSE, L2, mandala, and a simple classification accuracy); precise definitions are given in \Cref{sec:metrics}. 

\textbf{$(\mathbf{3\times64^2})$-D Image Data.} We then scale up the dimensionality to Tiny ImageNet \cite{2015LeTinyImageNet} using the public split (HuggingFace \cite{HuggingFaceTinyImageNet}). This dataset contains $64\times 64$ RGB images and its training set has 200 classes following WordNet synsets as in ImageNet. 

We trained an XS EDM2 model \cite{2024KarrasEDM2} as our main model for 22000 iterations. We defined a smaller XXS EDM2 model to use as the reference, and we trained it for only 5160 iterations. We used super-batch size $B=2048$ and mini batch-size $b=384$, effectively applying the same filtering ratio $f=0.8$ as in the 2D toy task. Further details can be found in \Cref{sec:supp-edm2}.
Evaluation uses FID \cite{2017HeuselFID}, FD-DINOv2 \cite{2024OquabDINOv2,2023SteinFD}, and top-1/top-5 accuracy of samples against a pretrained Tiny ImageNet Swin-L classifier \cite{2022HuynhTinySwinClassifier}; details are given in \Cref{sec:metrics}. 


\section{Results}

We report data-efficiency (metrics versus number of backpropagated examples) and time-efficiency (metrics versus wall-clock time, including selection overhead) as defined in \Cref{sec:experiments}. More results can be found in \Cref{sec:supp-results}.

\textbf{2-D Synthetic Data.} In the time-limited scenario, the 2D tree task results indicate that Early AJEST can be as time-efficient as random data selection and the baseline trained with no data selection. Despite this, Random may still be preferable due to its low algorithmic complexity and memory requirements. Results also show that Full AJEST is too time-consuming compared to other methods. 

In the data-limited scenario, full AJEST performs on a similar level to random data selection, but its added complexity and time overhead makes it less preferable. Early AJEST sacrifices data-efficiency in favour of time-efficiency, but it still presents advantages over the pure autoguidance approach: it reaches slightly better results, while using less examples for backpropagation. This begets the question whether Early AJEST has the potential to unlock better performance on tasks with higher complexity and dimensionality, such as image generation.

Additional results and experimentation with other AJEST variants can be found in \Cref{sec:supp-results-toy}.

\begin{table}[t]
  \fontsize{5.5pt}{7pt}\selectfont
  \centering
  \setlength{\tabcolsep}{2pt} 
  \renewcommand{\arraystretch}{1.2} 
  \caption{Evaluation metrics with guidance on the 2D task. Results are averaged over 5 runs with different random seeds. Standard deviation is used to indicate uncertainty. Yellow-filled cells contain the best scores. Undistinguishable results are marked in bold.}
  \label{tab:toy}
  \begin{tabular}{c l c c c c}
    \toprule
    \multirow{2}{*}{\textbf{Comparison}} & \multirow{2}{*}{\textbf{Method}} & \multicolumn{2}{c}{\textbf{Average L2 Distance}} & \multirow{2}{*}{\textbf{Mandala score}} & \multirow{2}{*}{\textbf{Classification score}} \\
    \cmidrule(lr){3-4}
    & & \textbf{Full Tree} & \textbf{External Branches} & & \\
    \midrule
    \multirow{4}{*}{\textbf{\begin{tabular}{c} Same \\ time \\ budget \end{tabular}}}
    & \cellcolor{gray!10} Baseline      & \cellcolor{gray!10} 0.246 $\pm$ 0.297 & \cellcolor{gray!10} 0.613 $\pm$ 0.210 & \cellcolor{gray!10} \textbf{0.73 $\pm$ 0.11} & \cellcolor{yellow!30}\textbf{0.94 $\pm$ 0.01} \\
    & Early AJEST   & 0.110 $\pm$ 0.019 & \textbf{0.516 $\pm$ 0.019} & \cellcolor{yellow!30}\textbf{0.75 $\pm$ 0.08} & \textbf{0.93 $\pm$ 0.02} \\
    & \cellcolor{gray!10} Full AJEST    & \cellcolor{gray!10} 0.118 $\pm$ 0.025 & \cellcolor{gray!10} \textbf{0.514 $\pm$ 0.020} & \cellcolor{gray!10} 0.53 $\pm$ 0.12 & \cellcolor{gray!10} 0.85 $\pm$ 0.04 \\
    & Random        & \cellcolor{yellow!30}\textbf{0.101 $\pm$ 0.006} & \cellcolor{yellow!30}\textbf{0.508 $\pm$ 0.009} & \textbf{0.71 $\pm$ 0.08} & \textbf{0.93 $\pm$ 0.01} \\
    \midrule
    \multirow{4}{*}{\textbf{\begin{tabular}{c}Same \\ data \\ budget\end{tabular}}}
    & \cellcolor{gray!10} Baseline      & \cellcolor{gray!10} 0.156 $\pm$ 0.053 & \cellcolor{gray!10} 0.543 $\pm$ 0.036 & \cellcolor{gray!10} 0.53 $\pm$ 0.15 & \cellcolor{gray!10} 0.83 $\pm$ 0.09 \\
    & Early AJEST   & 0.120 $\pm$ 0.029 & \textbf{0.515 $\pm$ 0.030} & 0.58 $\pm$ 0.14 & 0.86 $\pm$ 0.06 \\
    & \cellcolor{gray!10} Full AJEST    & \cellcolor{gray!10} \textbf{0.102 $\pm$ 0.006} & \cellcolor{yellow!30}\textbf{0.507 $\pm$ 0.009} & \cellcolor{gray!10} \textbf{0.69 $\pm$ 0.10} & \cellcolor{gray!10} \textbf{0.92 $\pm$ 0.02} \\
    & Random        & \cellcolor{yellow!30}\textbf{0.101 $\pm$ 0.005} & \textbf{0.508 $\pm$ 0.005} & \cellcolor{yellow!30}\textbf{0.72 $\pm$ 0.07} & \cellcolor{yellow!30}\textbf{0.93 $\pm$ 0.01} \\
    \bottomrule
  \end{tabular}
  \vspace{-.3cm}
\end{table}

\textbf{$(\mathbf{3\times64^2})$-D Image Data.} Our Tiny ImageNet experiment shows further evidence that the time overhead of Early AJEST is negligible: it reaches results almost as good as autoguidance on this fix time budget. AJEST proves to be inefficient from a time-based perspective for images too. 

\begin{figure}[t]
  \centerline{\includegraphics[width=0.42\textwidth]{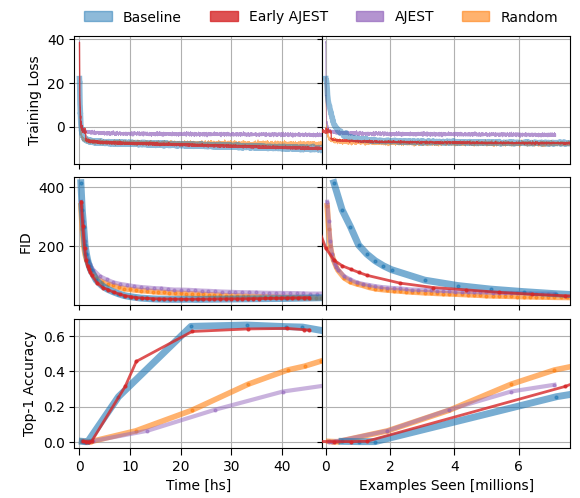}}
  \caption{Training loss and validation metrics on Tiny ImageNet for EMA=0.10 and guidance weight $\alpha$=2.2. \vale{Protocol/metrics in Setup.} Curves are shown versus number of backpropagated examples (data-efficiency) and wall-clock time (time-efficiency); error bars omitted for clarity.}
  \label{fig:images}
\end{figure}

Same as in the 2D task, Early AJEST achieved similar results to autoguidance on the fix data budget images scenario. It led to slightly better results on FID, FD-DINOv2 and Top-5 accuracy (Table \ref{tab:images_mean}), especially at very low data budgets (Figure \ref{fig:images}). However, random data selection presented better perceptual metrics compared to both Early AJEST and Baseline. The computational overhead of Early AJEST makes random selection preferable. 

Additional results including more generated images and other training and validation curves with all evaluated metrics can be found in \Cref{sec:supp-results-images}.
%

\noindent


\vspace{-3mm}\vspace{-.2cm}\begin{table}[t]
  \fontsize{5.5pt}{7pt}\selectfont
  \centering
  \setlength{\tabcolsep}{3pt} 
  \renewcommand{\arraystretch}{1.15}
  \caption{Evaluation metrics on Tiny ImageNet, averaging guided single-run results from EMA=0.05 and EMA=0.10 with $\alpha$=1.7 and $\alpha$=2.2 respectively. Yellow-filled cells contain the best scores.
  Close results are highlighted in bold.}
  \label{tab:images_mean}
  \begin{tabular}{
    p{1.2cm}  
    >{\centering\arraybackslash}p{1.4cm}  
    >{\centering\arraybackslash}p{1.0cm}  
    >{\centering\arraybackslash}p{1.2cm}  
    >{\centering\arraybackslash}p{1.0cm}  
    >{\centering\arraybackslash}p{1.0cm}  
  }
    \toprule
    \multirow{2}{*}{\textbf{Comparison}} & \multicolumn{1}{c}{\multirow{2}{*}{\textbf{Method}}} & \multicolumn{2}{c}{\textbf{Perceptual metrics}} & \multicolumn{2}{c}{\textbf{Classification-based metrics}} \\
    \cmidrule(lr){3-4} \cmidrule(lr){5-6}
    & & \textbf{FID} & \textbf{FD-DINOv2} & \textbf{Top-1} & \textbf{Top-5} \\
    \midrule
    \multirow{4}{*}{\textbf{\begin{tabular}{c}Same \\ time \\ budget\end{tabular}}}
    & \cellcolor{gray!10} Baseline     & \cellcolor{gray!10} 30.6 & \cellcolor{yellow!30}\textbf{624} & \cellcolor{yellow!30}\textbf{59.8} & \cellcolor{yellow!30}\textbf{80.6} \\
    & Early AJEST                     & \textbf{31.0} & \textbf{628} & \textbf{58.8} & \textbf{79.8} \\
    & \cellcolor{gray!10} AJEST       & \cellcolor{gray!10} 41.0 & \cellcolor{gray!10} 949 & \cellcolor{gray!10} 31.8 & \cellcolor{gray!10} 54.6 \\
    & Random                          & \cellcolor{yellow!30}\textbf{27.5} & 699 & 45.4 & 68.8 \\
    \midrule
    \multirow{4}{*}{\textbf{\begin{tabular}{c}Same \\ data \\ budget\end{tabular}}}
    & \cellcolor{gray!10} Baseline     & \cellcolor{gray!10} 40.4 & \cellcolor{gray!10} 934 & \cellcolor{yellow!30}\textbf{61.8} & \cellcolor{gray!10} \textbf{79.6} \\
    & Early AJEST                     & \textbf{34.7} & 849 & \textbf{60.7} & \cellcolor{yellow!30}\textbf{80.5} \\
    & \cellcolor{gray!10} AJEST       & \cellcolor{gray!10} 41.0 & \cellcolor{gray!10} 949 & \cellcolor{gray!10} 31.8 & \cellcolor{gray!10} 54.6 \\
    & Random                          & \cellcolor{yellow!30}\textbf{29.2} & \cellcolor{yellow!30}\textbf{737} & 40.9 & 65.0 \\
    \bottomrule
  \end{tabular}
  \vspace{-.4cm}
\end{table}

\section{Discussion and limitations}
\label{sec:limitations}

Our study examines online data curation for diffusion models when combined with autoguidance. Two consistent observations emerge across tasks.

First, autoguidance is a strong and reliable lever for improving quality while preserving diversity. It sets a robust baseline in our experiments that is difficult to surpass with data selection alone. Early AJEST---using the guide as the JEST reference only in the beginning of training---can yield modest data-efficiency gains and reach time-to-accuracy performance close to autoguidance on both the 2D and images task. However, the dominant improvements in coverage and fidelity are driven by autoguidance. 

\begin{figure}
    \centering
    \includegraphics[width=1.\linewidth]{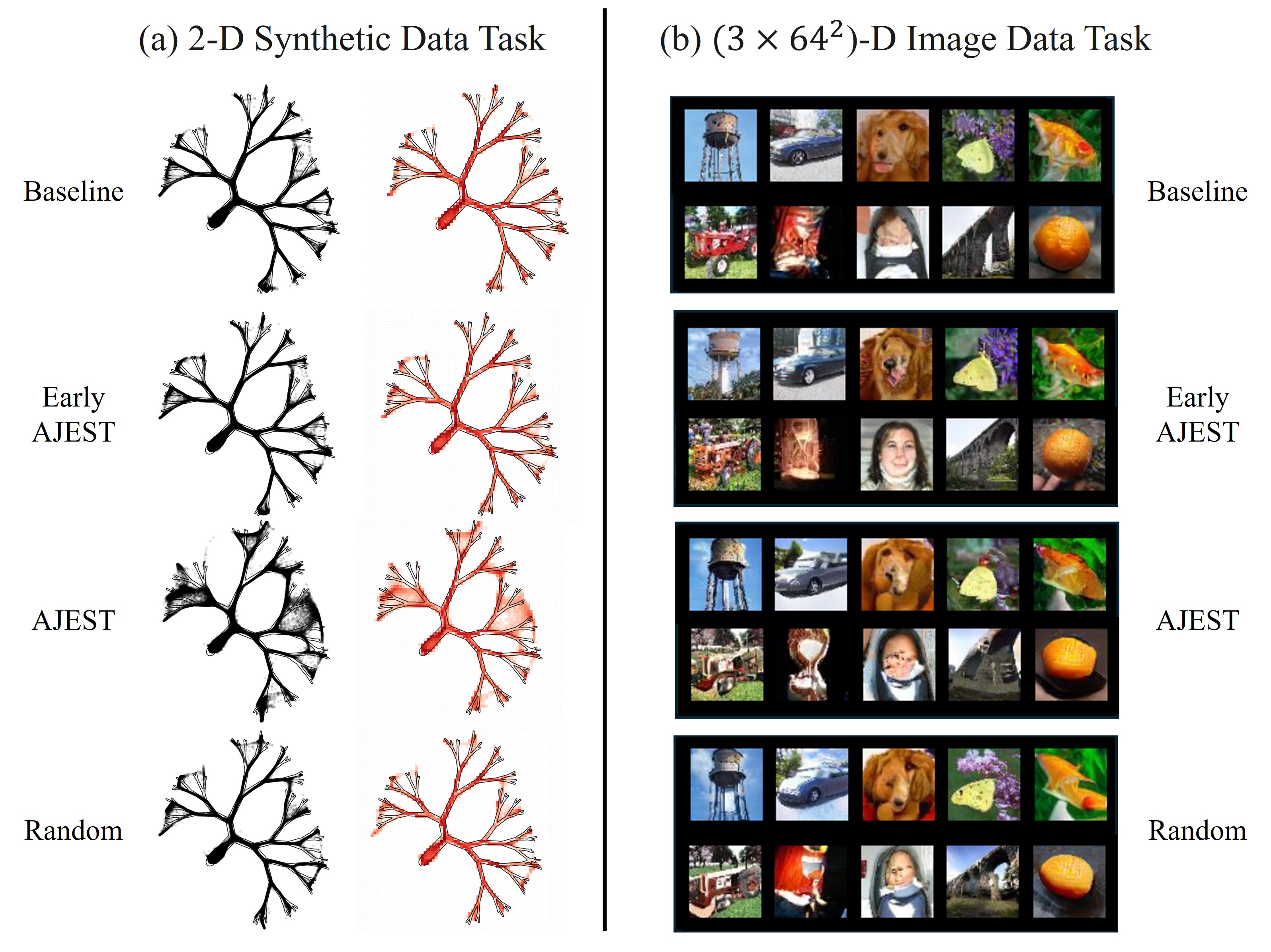}
    \caption{Qualitative results for fixed time budget on the (a) synthetic 2-D and (b) image generation tasks. For best viewing and fixed data budget results, please see full page figures in \Cref{sec:supp-results}.}
    \label{fig:qual-results}
    \vspace{-.5cm}
\end{figure}

Second, accounting for overhead matters. Full AJEST introduces significant runtime overhead and is consistently unattractive in any time-limited scenario. Early AJEST keeps the overhead negligible in our setup and can therefore be competitive when time is the primary constraint, but its advantages shrink under a fixed-data budget where autoguidance already performs strongly. Random subsetting is frequently competitive because of its near-zero complexity and memory footprint.

The following limitations should be considered when intepreting the results. \hbox{(i) Scope:} we evaluate on a controlled 2-D task and Tiny ImageNet at 64$\times$64; larger resolutions, datasets, and conditionings may alter the trade-offs. \hbox{(ii) Metrics:} while we report FD-DINOv2 and top-1/top-5 as more rigorous alternatives to FID, definitive perceptual assessment remains challenging. \hbox{(iii) Design choices:} we use only one reference model, one JEST parameterization and a simple early/always selection schedule; alternative parameters, chunking, and triggers could change outcomes. \hbox{(iv) Data selection:} we deploy AJEST, Early AJEST and a random uniform selection baseline; other methods, as those that use small proxy models, may offer improved synergy with autoguidance and diffusion models. \hbox{(v) Hyperparameters:} EMA and guidance weight were not exhaustively tuned; broader sweeps may shift the relative ranking at the margins. \hbox{(vi) Engineering:} results reflect a single hardware/software stack; absolute runtimes and overheads may vary.

Taken together, these results suggest a practical recipe: rely on autoguidance as the main driver of sample quality and diversity, and layer Early AJEST when training-time is scarce or when modest gains in data-efficiency are valuable relative to its small additional complexity. More aggressive selection schedules or heavier scoring generally do not justify their overhead in the setting explored here.

\section{Conclusions}
We evaluated Autoguided JEST (AJEST) for diffusion models on a 2-D toy task and Tiny ImageNet at $3\times64^2$ under equal time and data budgets. Autoguidance consistently improved quality and diversity and serves as a strong baseline. Early AJEST provided small but measurable data-efficiency gains with a minimal time overhead. In practice, we recommend autoguidance as the default, with Early AJEST added when training-time is constrained or when minor data-efficiency gains justify the added machinery. Future work includes exploring richer selection scores and schedules, larger and higher-resolution datasets, and broader metric suites.

\section*{Acknowledgments}

This work was supported by the UKRI EPSRC Centre for Doctoral Training in Applied Photonics [EP/S022821/1]. The main author's PhD scholarship had funding from University of Glasgow and Dotphoton A.G.
{
    \small
    \bibliographystyle{config/ieeenat_fullname}
    \bibliography{biblio}

@String(CVPR= {IEEE Conf. Comput. Vis. Pattern Recog.})

@String(ICCV= {Int. Conf. Comput. Vis.})

@String(ECCV= {Eur. Conf. Comput. Vis.})

@String(NIPS= {Adv. Neural Inform. Process. Syst.})

@String(CVPR  = {CVPR})

@String(ICCV  = {ICCV})

@String(ECCV  = {ECCV})

@String(NIPS  = {NeurIPS})

@misc{2024UdandaraoACID,
      title={Active Data Curation Effectively Distills Large-Scale Multimodal Models}, 
      author={Vishaal Udandarao and Nikhil Parthasarathy and Muhammad Ferjad Naeem and Talfan Evans and Samuel Albanie and Federico Tombari and Yongqin Xian and Alessio Tonioni and Olivier J. Hénaff},
      year={2025},
      eprint={2411.18674},
      archivePrefix={arXiv},
      primaryClass={cs.CV},
      url={https://arxiv.org/abs/2411.18674}, 
      note = {Accepted for the IEEE/CVF Conference on Computer Vision and Pattern Recognition 2025 to be hosted in June 2025.}
}

@inproceedings{2024KarrasAutoGuidance,
 author = {Karras, Tero and Aittala, Miika and Kynk\"{a}\"{a}nniemi, Tuomas and Lehtinen, Jaakko and Aila, Timo and Laine, Samuli},
 booktitle = {Advances in Neural Information Processing Systems},
 editor = {A. Globerson and L. Mackey and D. Belgrave and A. Fan and U. Paquet and J. Tomczak and C. Zhang},
 pages = {52996--53021},
 publisher = {Curran Associates, Inc.},
 title = {Guiding a Diffusion Model with a Bad Version of Itself},
 url = {https://proceedings.neurips.cc/paper_files/paper/2024/file/5ee7ed60a7e8169012224dec5fe0d27f-Paper-Conference.pdf},
 volume = {37},
 year = {2024}
}

@inproceedings{2022KarrasEDM,
author = {Karras, Tero and Aittala, Miika and Laine, Samuli and Aila, Timo},
title = {Elucidating the design space of diffusion-based generative models},
year = {2022},
isbn = {9781713871088},
publisher = {Curran Associates Inc.},
address = {Red Hook, NY, USA},
booktitle = {Proceedings of the 36th International Conference on Neural Information Processing Systems},
articleno = {1926},
numpages = {13},
location = {New Orleans, LA, USA},
series = {NIPS '22}
}

@InProceedings{2024KarrasEDM2,
    author    = {Karras, Tero and Aittala, Miika and Lehtinen, Jaakko and Hellsten, Janne and Aila, Timo and Laine, Samuli},
    title     = {Analyzing and Improving the Training Dynamics of Diffusion Models},
    booktitle = {Proceedings of the IEEE/CVF Conference on Computer Vision and Pattern Recognition (CVPR)},
    month     = {June},
    year      = {2024},
    pages     = {24174-24184}
}

@inproceedings{2024EvansJEST,
    title={Data curation via joint example selection further accelerates multimodal learning},
    author={Talfan Evans and Nikhil Parthasarathy and Hamza Merzic and Olivier J Henaff},
    booktitle={The Thirty-eight Conference on Neural Information Processing Systems Datasets and Benchmarks Track},
    year={2024},
    url={https://openreview.net/forum?id=xqpkzMfmQ5}
}

@misc{2024BriqPruningDiffusion,
      title={Data Pruning in Generative Diffusion Models}, 
      author={Rania Briq and Jiangtao Wang and Stefan Kesselheim},
      year={2025},
      eprint={2411.12523},
      archivePrefix={arXiv},
      primaryClass={cs.LG},
      url={https://arxiv.org/abs/2411.12523}, 
}

@techreport{2015LeTinyImageNet,
  author       = {Ya Le and Xuan S. Yang}, 
  title        = {Tiny ImageNet Classification with Convolutional Neural Networks}, 
  institution  = {Stanford University, CS231n: Convolutional Neural Networks for Visual Recognition},
  type         = {Course Project Report},
  number       = {Winter Quarter 2015},
  year         = {2015},
  note         = {Unpublished student project, available online},
  url          = {http://vision.stanford.edu/teaching/cs231n/reports/2015/pdfs/yle_project.pdf},
}

@misc{HuggingFaceTinyImageNet,
  title = {Tiny ImageNet (zh-plus version) on Hugging Face},
  author = {zh-plus},
  year = {2025},
  howpublished = {\url{https://huggingface.co/datasets/zh-plus/tiny-imagenet}},
  note = {Accessed: 2025-08-19}
}

@misc{KaggleTinyImageNet,
    title = {Tiny ImageNet (mnmoustafa version) on Kaggle},
    author = {mnmoustafa and Mohammed Ali},
    year = {2017},
    howpublished = {\url{https://kaggle.com/competitions/tiny-imagenet}},
    note = {Accessed: 2025-08-24}
}

@INPROCEEDINGS{2009DengImageNet,
  author={Deng, Jia and Dong, Wei and Socher, Richard and Li, Li-Jia and Kai Li and Li Fei-Fei},
  booktitle={2009 IEEE Conference on Computer Vision and Pattern Recognition}, 
  title={ImageNet: A large-scale hierarchical image database}, 
  year={2009},
  volume={},
  number={},
  pages={248-255},
  doi={10.1109/CVPR.2009.5206848}
}

@misc{2022HuynhTinySwinClassifier,
      title={Vision Transformers in 2022: An Update on Tiny ImageNet}, 
      author={Ethan Huynh},
      year={2022},
      eprint={2205.10660},
      archivePrefix={arXiv},
      primaryClass={cs.CV}
}

@inproceedings{2021PaulEL2NGradN,
 author = {Paul, Mansheej and Ganguli, Surya and Dziugaite, Gintare Karolina},
 booktitle = {Advances in Neural Information Processing Systems},
 editor = {M. Ranzato and A. Beygelzimer and Y. Dauphin and P.S. Liang and J. Wortman Vaughan},
 pages = {20596--20607},
 publisher = {Curran Associates, Inc.},
 title = {Deep Learning on a Data Diet: Finding Important Examples Early in Training},
 url = {https://proceedings.neurips.cc/paper_files/paper/2021/file/ac56f8fe9eea3e4a365f29f0f1957c55-Paper.pdf},
 volume = {34},
 year = {2021}
}

@article{
oala2024dmlr,
title={{DMLR}: Data-centric Machine Learning Research - Past, Present and Future},
author={Luis Oala and Manil Maskey and Lilith Bat-Leah and Alicia Parrish and Nezihe Merve G{\"u}rel and Tzu-Sheng Kuo and Yang Liu and Rotem Dror and Danilo Brajovic and Xiaozhe Yao and Max Bartolo and William A Gaviria Rojas and Ryan Hileman and Rainier Aliment and Michael W. Mahoney and Meg Risdal and Matthew Lease and Wojciech Samek and Debojyoti Dutta and Curtis G Northcutt and Cody Coleman and Braden Hancock and Bernard Koch and Girmaw Abebe Tadesse and Bojan Karla{\v{s}} and Ahmed Alaa and Adji Bousso Dieng and Natasha Noy and Vijay Janapa Reddi and James Zou and Praveen Paritosh and Mihaela van der Schaar and Kurt Bollacker and Lora Aroyo and Ce Zhang and Joaquin Vanschoren and Isabelle Guyon and Peter Mattson},
journal={Journal of Data-centric Machine Learning Research},
year={2024},
url={https://openreview.net/forum?id=2kpu78QdeE}
}

@article{BISCHL2025101317,
title = {OpenML: Insights from 10 years and more than a thousand papers},
journal = {Patterns},
volume = {6},
number = {7},
pages = {101317},
year = {2025},
issn = {2666-3899},
doi = {https://doi.org/10.1016/j.patter.2025.101317},
url = {https://www.sciencedirect.com/science/article/pii/S2666389925001655},
author = {Bernd Bischl and Giuseppe Casalicchio and Taniya Das and Matthias Feurer and Sebastian Fischer and Pieter Gijsbers and Subhaditya Mukherjee and Andreas C. Müller and László Németh and Luis Oala and Lennart Purucker and Sahithya Ravi and Jan N. {van Rijn} and Prabhant Singh and Joaquin Vanschoren and Jos {van der Velde} and Marcel Wever}
}

@inproceedings{2023TanMoSo,
author = {Tan, Haoru and Wu, Sitong and Du, Fei and Chen, Yukang and Wang, Zhibin and Wang, Fan and Qi, Xiaojuan},
title = {Data pruning via moving-one-sample-out},
year = {2023},
publisher = {Curran Associates Inc.},
address = {Red Hook, NY, USA},
booktitle = {Proceedings of the 37th International Conference on Neural Information Processing Systems},
articleno = {803},
numpages = {12},
location = {New Orleans, LA, USA},
series = {NIPS '23}
}

@inproceedings{2018SenerSavarese,
title={Active Learning for Convolutional Neural Networks: A Core-Set Approach},
author={Ozan Sener and Silvio Savarese},
booktitle={International Conference on Learning Representations},
year={2018},
url={https://openreview.net/forum?id=H1aIuk-RW},
}

@inproceedings{2020AgarwalContextual,
author = {Agarwal, Sharat and Arora, Himanshu and Anand, Saket and Arora, Chetan},
title = {Contextual Diversity for Active Learning},
year = {2020},
isbn = {978-3-030-58516-7},
publisher = {Springer-Verlag},
address = {Berlin, Heidelberg},
url = {https://doi.org/10.1007/978-3-030-58517-4_9},
doi = {10.1007/978-3-030-58517-4_9},
booktitle = {Computer Vision – ECCV 2020: 16th European Conference, Glasgow, UK, August 23–28, 2020, Proceedings, Part XVI},
pages = {137–153},
numpages = {17},
location = {Glasgow, United Kingdom}
}

@inproceedings{2023XiaModerate,
title={Moderate Coreset: A Universal Method of Data Selection for Real-world Data-efficient Deep Learning},
author={Xiaobo Xia and Jiale Liu and Jun Yu and Xu Shen and Bo Han and Tongliang Liu},
booktitle={The Eleventh International Conference on Learning Representations },
year={2023},
url={https://openreview.net/forum?id=7D5EECbOaf9}
}

@InProceedings{2020MirzasoleimanCRAIG,
  title = 	 {Coresets for Data-efficient Training of Machine Learning Models},
  author =       {Mirzasoleiman, Baharan and Bilmes, Jeff and Leskovec, Jure},
  booktitle = 	 {Proceedings of the 37th International Conference on Machine Learning},
  pages = 	 {6950--6960},
  year = 	 {2020},
  editor = 	 {III, Hal Daumé and Singh, Aarti},
  volume = 	 {119},
  series = 	 {Proceedings of Machine Learning Research},
  month = 	 {13--18 Jul},
  publisher =    {PMLR},
  url = 	 {https://proceedings.mlr.press/v119/mirzasoleiman20a.html}
}

@InProceedings{2022PooladzandiAdaCore,
  title = 	 {Adaptive Second Order Coresets for Data-efficient Machine Learning},
  author =       {Pooladzandi, Omead and Davini, David and Mirzasoleiman, Baharan},
  booktitle = 	 {Proceedings of the 39th International Conference on Machine Learning},
  pages = 	 {17848--17869},
  year = 	 {2022},
  editor = 	 {Chaudhuri, Kamalika and Jegelka, Stefanie and Song, Le and Szepesvari, Csaba and Niu, Gang and Sabato, Sivan},
  volume = 	 {162},
  series = 	 {Proceedings of Machine Learning Research},
  month = 	 {17--23 Jul},
  publisher =    {PMLR},
  url = 	 {https://proceedings.mlr.press/v162/pooladzandi22a.html}
}

@InProceedings{2023ShinLCMat,
  title = 	 {Loss-Curvature Matching for Dataset Selection and Condensation},
  author =       {Shin, Seungjae and Bae, Heesun and Shin, Donghyeok and Joo, Weonyoung and Moon, Il-Chul},
  booktitle = 	 {Proceedings of The 26th International Conference on Artificial Intelligence and Statistics},
  pages = 	 {8606--8628},
  year = 	 {2023},
  editor = 	 {Ruiz, Francisco and Dy, Jennifer and van de Meent, Jan-Willem},
  volume = 	 {206},
  series = 	 {Proceedings of Machine Learning Research},
  month = 	 {25--27 Apr},
  publisher =    {PMLR},
  url = 	 {https://proceedings.mlr.press/v206/shin23a.html}
}

@inproceedings{2018TonevaForgetting,
title={An Empirical Study of Example Forgetting during Deep Neural Network Learning},
author={Mariya Toneva and Alessandro Sordoni and Remi Tachet des Combes and Adam Trischler and Yoshua Bengio and Geoffrey J. Gordon},
booktitle={International Conference on Learning Representations},
year={2019},
url={https://openreview.net/forum?id=BJlxm30cKm},
}

@inproceedings{2020FeldmanMemorization,
 author = {Feldman, Vitaly and Zhang, Chiyuan},
 booktitle = {Advances in Neural Information Processing Systems},
 editor = {H. Larochelle and M. Ranzato and R. Hadsell and M.F. Balcan and H. Lin},
 pages = {2881--2891},
 publisher = {Curran Associates, Inc.},
 title = {What Neural Networks Memorize and Why: Discovering the Long Tail via Influence Estimation},
 url = {https://proceedings.neurips.cc/paper_files/paper/2020/file/1e14bfe2714193e7af5abc64ecbd6b46-Paper.pdf},
 volume = {33},
 year = {2020}
}

@inproceedings{2020ColemanSVP,
title={Selection via Proxy: Efficient Data Selection for Deep Learning},
author={Cody Coleman and Christopher Yeh and Stephen Mussmann and Baharan Mirzasoleiman and Peter Bailis and Percy Liang and Jure Leskovec and Matei Zaharia},
booktitle={International Conference on Learning Representations},
year={2020},
url={https://openreview.net/forum?id=HJg2b0VYDr}
}

@inproceedings{2020AshAgarwalBADGE,
title={Deep Batch Active Learning by Diverse, Uncertain Gradient Lower Bounds},
author={Jordan T. Ash and Chicheng Zhang and Akshay Krishnamurthy and John Langford and Alekh Agarwal},
booktitle={International Conference on Learning Representations},
year={2020},
url={https://openreview.net/forum?id=ryghZJBKPS}
}

@inproceedings{2018HanCoTeaching,
 author = {Han, Bo and Yao, Quanming and Yu, Xingrui and Niu, Gang and Xu, Miao and Hu, Weihua and Tsang, Ivor and Sugiyama, Masashi},
 booktitle = {Advances in Neural Information Processing Systems},
 editor = {S. Bengio and H. Wallach and H. Larochelle and K. Grauman and N. Cesa-Bianchi and R. Garnett},
 pages = {},
 publisher = {Curran Associates, Inc.},
 title = {Co-teaching: Robust training of deep neural networks with extremely noisy labels},
 url = {https://proceedings.neurips.cc/paper_files/paper/2018/file/a19744e268754fb0148b017647355b7b-Paper.pdf},
 volume = {31},
 year = {2018}
}

@InProceedings{2018JiangMentorNet,
  title = 	 {{M}entor{N}et: Learning Data-Driven Curriculum for Very Deep Neural Networks on Corrupted Labels},
  author =       {Jiang, Lu and Zhou, Zhengyuan and Leung, Thomas and Li, Li-Jia and Fei-Fei, Li},
  booktitle = 	 {Proceedings of the 35th International Conference on Machine Learning},
  pages = 	 {2304--2313},
  year = 	 {2018},
  editor = 	 {Dy, Jennifer and Krause, Andreas},
  volume = 	 {80},
  series = 	 {Proceedings of Machine Learning Research},
  month = 	 {10--15 Jul},
  publisher =    {PMLR},
  url = 	 {https://proceedings.mlr.press/v80/jiang18c.html}
}

@InProceedings{2024GoyalAgnosticScaling,
    author    = {Goyal, Sachin and Maini, Pratyush and Lipton, Zachary C. and Raghunathan, Aditi and Kolter, J. Zico},
    title     = {Scaling Laws for Data Filtering-- Data Curation cannot be Compute Agnostic},
    booktitle = {Proceedings of the IEEE/CVF Conference on Computer Vision and Pattern Recognition (CVPR)},
    month     = {June},
    year      = {2024},
    pages     = {22702-22711}
}

@inproceedings{2022SorscherBeatingScalingCNNs,
 author = {Sorscher, Ben and Geirhos, Robert and Shekhar, Shashank and Ganguli, Surya and Morcos, Ari},
 booktitle = {Advances in Neural Information Processing Systems},
 editor = {S. Koyejo and S. Mohamed and A. Agarwal and D. Belgrave and K. Cho and A. Oh},
 pages = {19523--19536},
 publisher = {Curran Associates, Inc.},
 title = {Beyond neural scaling laws: beating power law scaling via data pruning},
 url = {https://proceedings.neurips.cc/paper_files/paper/2022/file/7b75da9b61eda40fa35453ee5d077df6-Paper-Conference.pdf},
 volume = {35},
 year = {2022}
}

@misc{2020KaplanScalingLanguage,
      title={Scaling Laws for Neural Language Models}, 
      author={Jared Kaplan and Sam McCandlish and Tom Henighan and Tom B. Brown and Benjamin Chess and Rewon Child and Scott Gray and Alec Radford and Jeffrey Wu and Dario Amodei},
      year={2020},
      eprint={2001.08361},
      archivePrefix={arXiv},
      primaryClass={cs.LG},
      url={https://arxiv.org/abs/2001.08361}, 
}

@InProceedings{2023ChertiScalingCLIP,
    author    = {Cherti, Mehdi and Beaumont, Romain and Wightman, Ross and Wortsman, Mitchell and Ilharco, Gabriel and Gordon, Cade and Schuhmann, Christoph and Schmidt, Ludwig and Jitsev, Jenia},
    title     = {Reproducible Scaling Laws for Contrastive Language-Image Learning},
    booktitle = {Proceedings of the IEEE/CVF Conference on Computer Vision and Pattern Recognition (CVPR)},
    month     = {June},
    year      = {2023},
    pages     = {2818-2829}
}

@article{2024OquabDINOv2,
    title={{DINO}v2: Learning Robust Visual Features without Supervision},
    author={Maxime Oquab and Timoth{\'e}e Darcet and Th{\'e}o Moutakanni and Huy V. Vo and Marc Szafraniec and Vasil Khalidov and Pierre Fernandez and Daniel HAZIZA and Francisco Massa and Alaaeldin El-Nouby and Mido Assran and Nicolas Ballas and Wojciech Galuba and Russell Howes and Po-Yao Huang and Shang-Wen Li and Ishan Misra and Michael Rabbat and Vasu Sharma and Gabriel Synnaeve and Hu Xu and Herve Jegou and Julien Mairal and Patrick Labatut and Armand Joulin and Piotr Bojanowski},
    journal={Transactions on Machine Learning Research},
    issn={2835-8856},
    year={2024},
    url={https://openreview.net/forum?id=a68SUt6zFt},
    note={Featured Certification}
}

@misc{2022HoCFG,
      title={Classifier-Free Diffusion Guidance}, 
      author={Jonathan Ho and Tim Salimans},
      year={2022},
      eprint={2207.12598},
      archivePrefix={arXiv},
      primaryClass={cs.LG},
      url={https://arxiv.org/abs/2207.12598}, 
}

@inproceedings{2017HeuselFID,
 author = {Heusel, Martin and Ramsauer, Hubert and Unterthiner, Thomas and Nessler, Bernhard and Hochreiter, Sepp},
 booktitle = {Advances in Neural Information Processing Systems},
 editor = {I. Guyon and U. Von Luxburg and S. Bengio and H. Wallach and R. Fergus and S. Vishwanathan and R. Garnett},
 pages = {},
 publisher = {Curran Associates, Inc.},
 title = {GANs Trained by a Two Time-Scale Update Rule Converge to a Local Nash Equilibrium},
 url = {https://proceedings.neurips.cc/paper_files/paper/2017/file/8a1d694707eb0fefe65871369074926d-Paper.pdf},
 volume = {30},
 year = {2017}
}

@inproceedings{2023SteinFD,
 author = {Stein, George and Cresswell, Jesse and Hosseinzadeh, Rasa and Sui, Yi and Ross, Brendan and Villecroze, Valentin and Liu, Zhaoyan and Caterini, Anthony L and Taylor, Eric and Loaiza-Ganem, Gabriel},
 booktitle = {Advances in Neural Information Processing Systems},
 editor = {A. Oh and T. Naumann and A. Globerson and K. Saenko and M. Hardt and S. Levine},
 pages = {3732--3784},
 publisher = {Curran Associates, Inc.},
 title = {Exposing flaws of generative model evaluation metrics and their unfair treatment of diffusion models},
 url = {https://proceedings.neurips.cc/paper_files/paper/2023/file/0bc795afae289ed465a65a3b4b1f4eb7-Paper-Conference.pdf},
 volume = {36},
 year = {2023}
}

@INPROCEEDINGS{2016SzegedyInceptionV3,
  author={Szegedy, Christian and Vanhoucke, Vincent and Ioffe, Sergey and Shlens, Jon and Wojna, Zbigniew},
  booktitle={2016 IEEE Conference on Computer Vision and Pattern Recognition (CVPR)}, 
  title={Rethinking the Inception Architecture for Computer Vision}, 
  year={2016},
  volume={},
  number={},
  pages={2818-2826},
  doi={10.1109/CVPR.2016.308}}

@INPROCEEDINGS{2021LiuSwin,
  author={Liu, Ze and Lin, Yutong and Cao, Yue and Hu, Han and Wei, Yixuan and Zhang, Zheng and Lin, Stephen and Guo, Baining},
  booktitle={2021 IEEE/CVF International Conference on Computer Vision (ICCV)}, 
  title={Swin Transformer: Hierarchical Vision Transformer using Shifted Windows}, 
  year={2021},
  volume={},
  number={},
  pages={9992-10002},
  doi={10.1109/ICCV48922.2021.00986}
}

@inproceedings{
    2024OkanovicRS2,
    title     = {Repeated Random Sampling for Minimizing the Time-to-Accuracy of Learning},
    author    = {Patrik Okanovic and Roger Waleffe and Vasilis Mageirakos and Konstantinos Nikolakakis and Amin Karbasi and Dionysios Kalogerias and Nezihe Merve G{\"u}rel and Theodoros Rekatsinas},
    booktitle = {The Twelfth International Conference on Learning Representations},
    year      = {2024},
    url       = {https://openreview.net/forum?id=JnRStoIuTe}
}
}

\clearpage
\setcounter{page}{1}
\maketitlesupplementary

\appendix
\crefname{appendix}{Appendix}{Appendices}
\Crefname{appendix}{Appendix}{Appendices}
\crefalias{section}{appendix}

\section{Implementation details}
\label{sec:implementation}

\subsection{JEST data selection}
\label{sec:supp-ajest}

Joint example selection (JEST) samples training examples based on a learnability score \cite{2024EvansJEST}. Assuming that datapoints with indices $i\in{1,...,B}$ are fed into the learner model and samples with indices $j\in{1,...,B}$ are fed into the reference model, the learnability score $s_{ij}$ is defined as...

\begin{equation}
    \label{eq:learnability}
    s_{ij}^{learn} = L_{i}^{L} - L^{R}_{j}
\end{equation}

Where $L^L$ is the loss evaluated by the learner model on the $i$th example of the batchand $L^R$ is the loss evaluated by the reference model on the $j$th example.

\begin{figure}[h]
  \centerline{\includegraphics[width=0.42\textwidth]{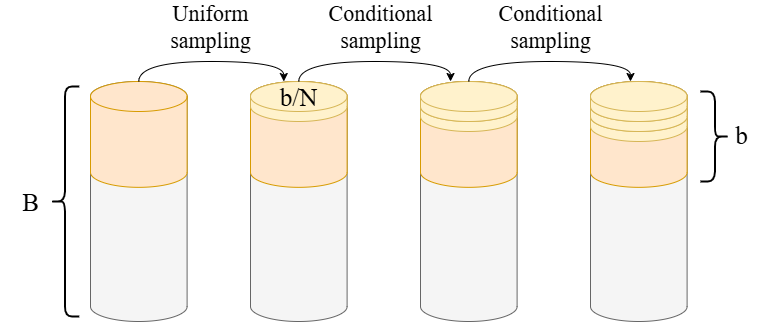}}
  \caption{Iterative sampling process for JEST data selection.}
  \label{fig:jest}
\end{figure}

The joint batch selection is done in an iterative process with $N$ steps by selecting $n$ chunks of size $\frac{b}{N}$ (Figure \ref{fig:jest}). The first chunk $C_0$ is populated sampling from a uniform probability distribution over all $B$ datapoints. For each successive chunk $C_n$, sampling is done on a conditional probability distribution over the $B$ datapoints given all $n\frac{b}{N}$ previously selected datapoints that define a $\mathcal{K} = C_1 \, \cup \, ... \, \cup \, C_n$ set. This conditional probability distribution is modeled as a softmax distribution that takes certain logits as input. The logits $\mathbf{z} = (z_1, ..., z_B)$ are calculated adding four vector terms of length $B$. 

\begin{enumerate}
    \item The $(s_{11}, ..., s_{BB})$ diagonal scores that feed the same datapoint to both the learner and the reference models.
    \item The sums of scores $(\sum_{k\in K} s_{k1}, ..., \sum_{k\in K} s_{kB})$ that result from only considering the selected datapoints fed into the learner model.
    \item The sums of scores $(\sum_{k\in K} s_{1k}, ..., \sum_{k\in K} s_{Bk})$ that result from only considering the selected datapoints fed into the reference model.
    \item A penalizing term whose elements are $-10^8$ for all selected datapoints in $K$ and 0 for unselected datapoints.
\end{enumerate}

Our code implementation follows quite closely the one published by Evans et al \cite{2024EvansJEST}. However, we observed that a softmax distribution applied directly to these $\mathbf{z}$ logits is highly unstable. For positive increasing scores, the exponential of the logits is unbounded and quickly causes a numerical overflow. We therefore applied a normalization factor to the logits before using them as input to the softmax function. We noticed that at any given JEST iteration with index $n$, the logits are calculated as the sum of $n.\frac{b}{N}$ terms. To bound the logits during the full iterative process, we simply divided them by that number, resulting in the following final expression:

\begin{equation}
    z_{l} = \frac{N}{nb} \; \bigg( s_{ll} + \sum_{k\in \mathcal{K}} s_{kl} + \sum_{k\in \mathcal{K}} s_{lk} - \alpha_l \bigg)
\end{equation}

\subsection{Autoguidance and hyperparameters}
\label{sec:supp-autoguidance}

By Karras et al.'s terminology \cite{2024KarrasAutoGuidance}, diffusion models are used to evolve any noisy sample $\mathbf{x}_{\text{initial}} \sim p(\mathbf{x};\sigma_{\text{max}})$ from a Gaussian distribution $\mathcal{N}(\mathbf{x}; \sigma^2\mathbf{I})$ with maximum noise level $\sigma_{\text{max}}$ to a sample matching the ground truth distribution $\mathbf{x}_{\text{final}} \sim p(\mathbf{x};0)$ with zero noise. The following probability flow ordinary differential equation needs to be solved:

\begin{equation}
    \label{eq:pf-ode}
    d\mathbf{x}_\sigma = - \sigma \; S(\mathbf{x}_\sigma ; \sigma) \; d\sigma
\end{equation}

\begin{equation}
    \label{eq:score}
    S(\mathbf{x}_\sigma ; \sigma) = \nabla_{\mathbf{x}_\sigma} \log{ p (\mathbf{x}_\sigma ; \sigma) }
\end{equation}

The diffusion model is trained to predict the score for every $\sigma \in [0, \sigma_{\text{max}}]$ and every example $\mathbf{x}_\sigma \sim p (\mathbf{x}_\sigma ; \sigma)$. A $\sigma = t$ noise schedule set by a time step $t$ can be used to obtain the final sample using an iterative approach \cite{2022KarrasEDM}.
Autoguidance improves the performance of diffusion models replacing the score from \cref{eq:pf-ode} with an interpolation of scores defined as in \cref{eq:autoguided-score}: 

\begin{equation}
    \label{eq:autoguided-score}
    S = \alpha S^{\text{main}} + (1-\alpha) S^{\text{ref}}
\end{equation}

The $S^{\text{main}}$ score is evaluated by the main model and the $S^{\text{ref}}$ score is evaluated by an auxiliary reference model. The two models share the same architecture and conditioning, but the reference model is assumed to be a weaker version of the main model: a smaller model that is trained for less iterations. This applies a corrective force that pushes samples on each step of the denoising process towards regions on which the reference and the main model disagree on their predictions. This guidance is beneficial because it acts on the assumption that both the reference and the main model will make similar mistakes.

The guidance weight $\alpha$ is a key hyperparameter in our experiments. Autoguidance influences the results for any $\alpha>1$. If set to $\alpha=1$, the main scores are retrieved; if set to $\alpha=0$, the reference scores are retrieved instead. Another key hyperparameter is the EMA length: the duration over which an Exponential Moving Average (EMA) on the models weights is calculated during training. 

Whenever available, we use $\alpha$ and EMA values directly extracted from Karras et al.'s publication \cite{2024KarrasAutoGuidance}. For the 2D tree task, we use their default value $\alpha=3$. On ImageNet-64 with EDM-XS, Karras et al report the best FID metric for $\alpha=1.7$ and EMA=0.045 and the best FD-DINOv2 metric for $\alpha=2.2$ and EMA=0.105. We train with default EMA=0.100 and EMA=0.050 values, so we adopt $\alpha=2.2$ for the first one and $\alpha=1.7$ for the second one.

\subsection{2D tree data generation task}
\label{sec:supp-2d}

\begin{figure}[b!]
  \centerline{\includegraphics[width=0.3\textwidth]{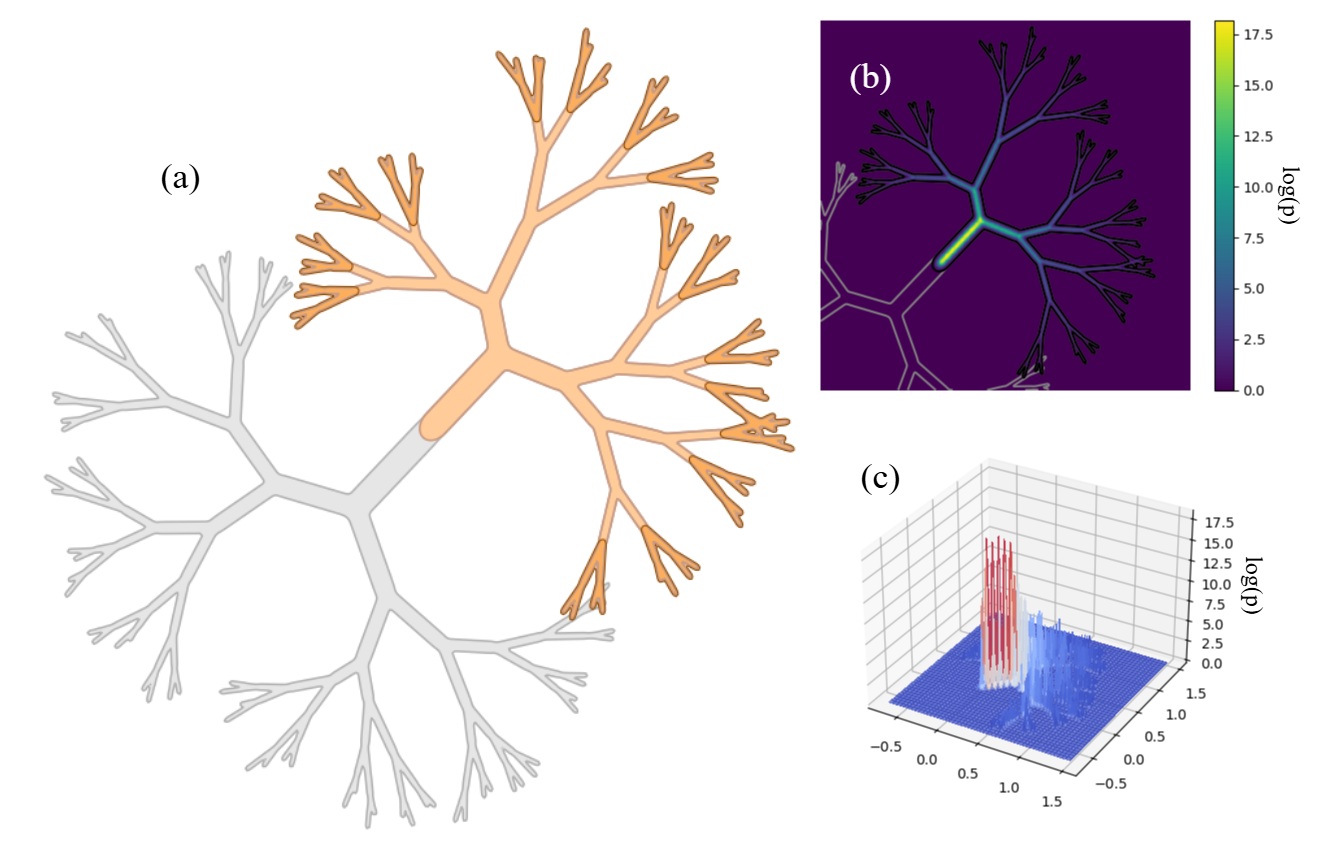}}
  \caption{Visualization of the 2D tree ground-truth distribution. (a) A contour of the 2D tree distribution, showing which areas are considered class A and its external branches. (b) Color map of the logarithm of the ground-truth probability distribution function $\log{p}$. (c) A mixture of Gaussians is used to model the tree distribution and its peaks are visible in the 3D plot of $\log{p}$.}
  \label{fig:tree-full}
\end{figure}

The ground truth distribution for this task is modeled with a mixture of multi-variate 2D Gaussians (see Figure \ref{fig:tree-full}). Each brach is created out of 8 Gaussians uniformly-distributed along its central line segment. The exact same parameters from Karras et al \cite{2024KarrasAutoGuidance} were used. The upper half of the tree is assigned to class A -the class we aim to sample from- and the lower half of the tree is assigned to class B. The tree is designed to have depth level 7, meaning a branch is split in two 7 times. To gain further resolution on our metrics and data analysis, we define as "external branches" all those branches with depth level 5 or more.

\noindent
\noindent

As described in \cite{2024KarrasAutoGuidance}, we built 2D diffusion models following a simple multi-layer perceptron architecture. The neural network's input consisted of a set of $(x, y)$ coordinates and the $\sigma$ noise level that correlates with the time step $t$ of the denoising process. Four hidden layers with the same number of neurons followed the input layer. The scalar output was trained to represent the logarithm of the unnormalized probability density. Differentiating this value, the model could also return the score function. Training was done via exact score-matching. The loss function was defined as the mean squared error between the score and the ground truth score scaled by a $\sigma^2$ factor (see \Cref{eq:loss} in \ref{sec:metrics}).

Following Karras et al. \cite{2024KarrasAutoGuidance}, we first trained a small reference model for only 512 iterations with hidden dimension 32. We then independently trained a larger main model with hidden dimension 64 for 4096 iterations. We trained on a single NVIDIA A5000 GPU. In this setup, a no-selection baseline could be trained for 4096 iterations in approximately 23 minutes; Full AJEST increased this to $\approx36$ minutes, while Early AJEST added only $\approx 38$ seconds over baseline.

We store models every 128 training iterations and we evaluate data and time efficiency at the latest point possible during training. For the equal wall-time scenario, this happens at the last epoch of Baseline, the fastest method; the closest matching checkpoints are collected from all other methods, resulting on an average training time of $(16.6\,\pm\,0.5)$ min. For the scenario with equal number of backpropagated examples, this happens at the last epoch of AJEST and Random, the methods with the highest number of filtered examples; closest matching checkpoints result on an average of $(3.3\,\pm\,0.2)$ million examples.

\subsection{EDM2 for Tiny ImageNet}
\label{sec:supp-edm2}

As Karras et al. \cite{2024KarrasAutoGuidance}, we used the EDM2 diffusion model architecture \cite{2024KarrasEDM2} for image generation. EDM2 diffusion models are based on a UNet network with 4 blocks on its decoder and encoder. The XS EDM2 model has 128 channels on its first UNet block, whereas the XXS EDM2 model desgined by us has 64 channels instead. We trained these models to solve a natural images generation task on Tiny ImageNet \cite{2015LeTinyImageNet} as found in HuggingFace \cite{HuggingFaceTinyImageNet}. The training dataset contains 500 $64\times 64$ RGB images for each of its 200 classes, adding up to a total of 1 million training examples. The categories correspond to synsets of the WordNet hierarchy, same as they do in the widely-used ImageNet dataset.

We know that Karras et al. \cite{2024KarrasAutoGuidance} trained the XS model with batch size 2048 until it had seen 2147.5 Mimg from ImageNet, a dataset containing 1,281,167 training examples. We therefore estimated we would need to train the XS model for 167.6 Mimg in Tiny ImageNet to achieve convergence, assuming linear scaling on the training dataset size out of simplicity. Karras et al. \cite{2024KarrasAutoGuidance} used a smaller model trained for 1/16th training iterations as their reference, so we decided we could use an XXS EDM2 model trained for 5160 iterations instead of the 81845 iterations required to go through 167.6 Mimg at batch size 2048.

We executed parallel-data distributed training on 2 NVIDIA A5000 GPUs. However, our dual GPU system could only achieve a speed of \hbox{8.2 sec/epoch} or \hbox{1.1 hs/Mimg}, forcing the convergence of the XS model to require 7.7 days. As we had seen that the best results in the 2D task could only be achieved by early data selection, we decided to stop training the XS main model at 48 hs and 22000 iterations. However, we kept the reference model trained for 5160 iterations. In consequence, our reference model was trained for a 1/4th of the main model's total iterations, 4 times more what Karras et al do.

We converted the 8-bits Tiny ImageNet images into $[-1,1]$ float images, and applied no other normalization besides the input preconditioning from the EDM2 model. We used Adam optimizer and we adopted Karras et al's EDM2 learning rate schedule \cite{2024KarrasEDM2}. However, to train with data selection we follow Okanovik et al's recommendation \cite{2024OkanovicRS2} on how to adapt the learning rate schedule to decay as fast as it would when training on the entire dataset (see Figure 1 from \cite{2024KarrasEDM2}).

We stored models every 500 training iterations. We followed the same methodology as in the 2D tree task to select a fix data and time budget for our analysis. For the equal wall-time scenario, this corresponded to $(47.73\,\pm\,0.07)$ hs; for the scenario with equal number of backpropagated examples, $(7.2\,\pm\,0.2)$ Mimg.


\section{Additional results}
\label{sec:supp-results}

\subsection{2D toy example with all metrics evaluated}
\label{sec:supp-results-toy}

\begin{figure*}
  \centerline{\includegraphics[width=0.9\textwidth]{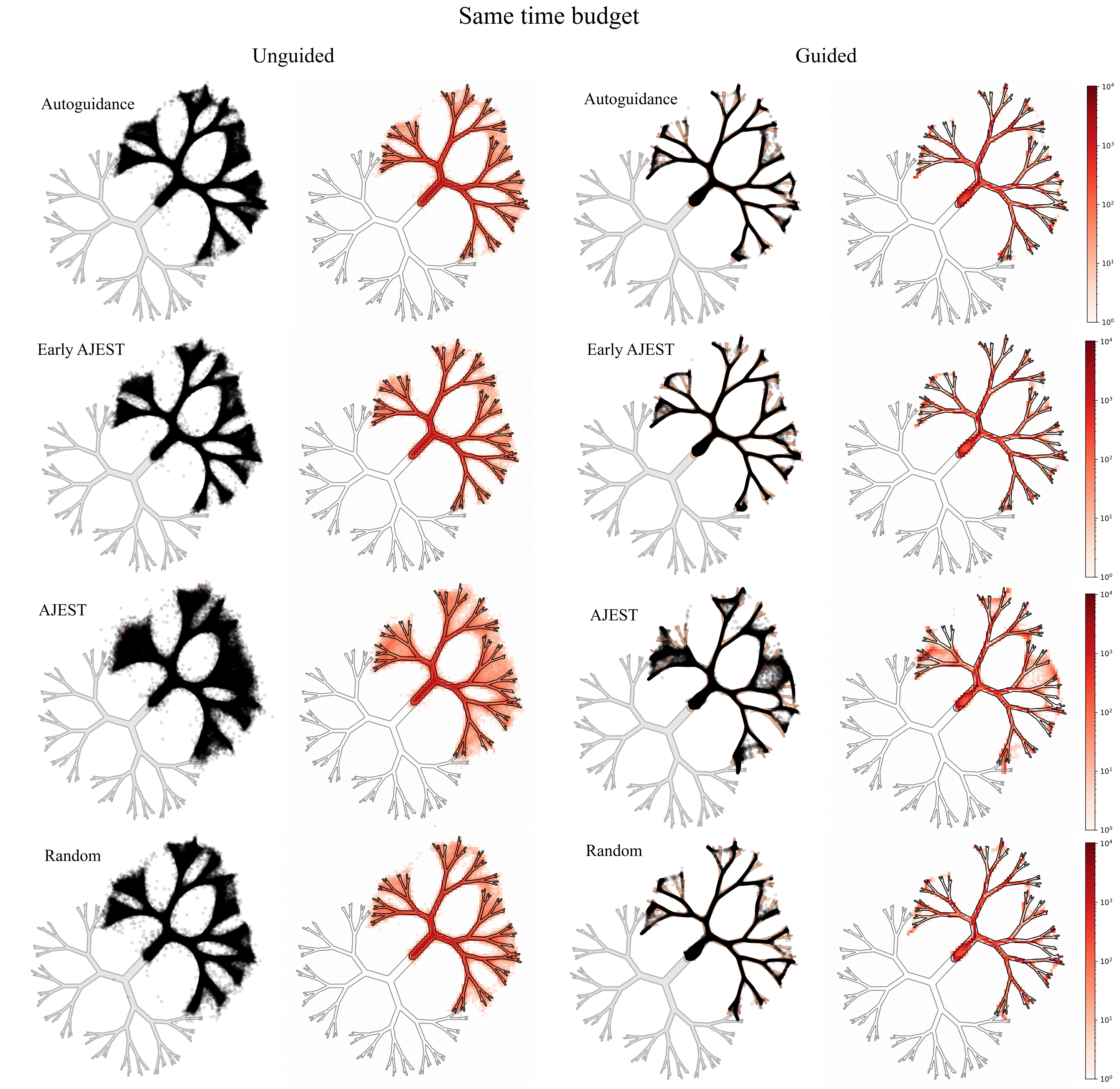}}
  \caption{Distribution of generated points on the 2D tree task for the fix time budget of (16.6 $\pm$ 0.5) min.}
  \label{fig:mandalas-time}
\end{figure*}

\begin{figure*}
  \centerline{\includegraphics[width=0.9\textwidth]{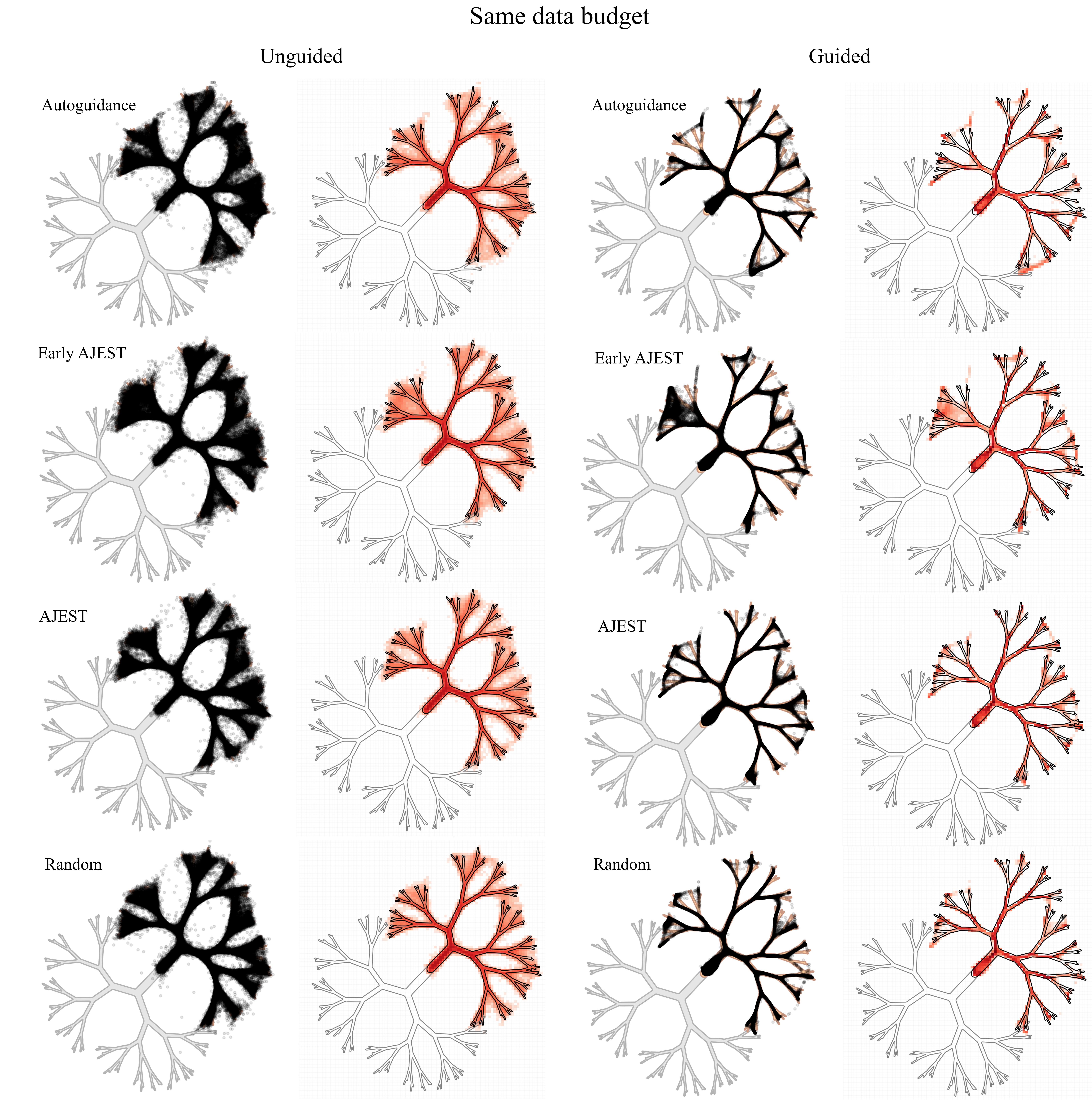}}
  \caption{Distribution of generated points on the 2D tree task for the fix data budget of (3.3 $\pm$ 0.2) million training examples.}
  \label{fig:mandalas-data}
\end{figure*}

Results for the complete set of metrics evaluated both with guidance and no guidance at equal wall time and equal number of backpropagated examples are presented on Table \ref{tab:toy-efficiency-full}. We also include visual representations of the distribution of generated points in Figures \ref{fig:mandalas-time} and \ref{fig:mandalas-data} for the fixed time and fixed data budget scenarios, both with guidance and without it.

Unguided models reach better performance on the average L2 metric in external branches, but this is in clear conflict with the perceptual quality assessment that can be done on the generated points distributions. As Karras et al. state, guided models generate points closer to the high-density regions of the distribution, avoiding regions between branches where the ground-truth probability density is low. This is what originally led us to develop the classification and mandala scores (see \Cref{sec:metrics}), a set of metrics that better reflects autoguidance advantages and rewards the desired behaviour of diffusion models in the 2D toy task.

\begin{table*}[h!]
  \tiny
  \centering
  \renewcommand{\arraystretch}{1.2}
  \caption{All evaluation metrics applied to the 2D tree task. Results are averaged over 5 runs with different random seeds. Standard deviation is used to indicate uncertainty. Yellow-filled cells indicate the best scores for each metric: the lowest for the loss and L2 distance, and the highest for classification-based scores. Undistinguishable results according to the uncertainty are marked in bold.}
  \label{tab:toy-efficiency-full}
  \resizebox{\textwidth}{!}{%
  \begin{tabular}{ll
                  cc
                  cc
                  cc
                  cc
                  cc}
    \toprule
    \multirow{3}{*}{\textbf{Comparison}} & \multirow{3}{*}{\textbf{Method}} 
    & \multicolumn{2}{c}{\textbf{Average Loss}} 
    & \multicolumn{4}{c}{\textbf{Average L2 Distance}} 
    & \multicolumn{2}{c}{\textbf{Mandala score}} 
    & \multicolumn{2}{c}{\textbf{Classification score}} \\
    
    \cmidrule(lr){3-4} \cmidrule(lr){5-8} \cmidrule(lr){9-10} \cmidrule(lr){11-12}
    
    & & \textbf{Full Tree} & \textbf{External Branches} 
      & \multicolumn{2}{c}{\textbf{Full Tree}} & \multicolumn{2}{c}{\textbf{External Branches}} 
      & \textbf{Unguided} & \textbf{Guided} 
      & \textbf{Unguided} & \textbf{Guided} \\
    
    \cmidrule(lr){5-6} \cmidrule(lr){7-8}

    & & & & \textbf{Unguided} & \textbf{Guided} & \textbf{Unguided} & \textbf{Guided} & & & & \\
    
    \midrule
    \multirow{4}{*}{\textbf{\begin{tabular}{c} Same \\ time \\ budget \end{tabular}}} 
    & \cellcolor{gray!10} Baseline & \cellcolor{yellow!30}\textbf{0.011 $\pm$ 0.003} & \cellcolor{yellow!30}\textbf{0.227 $\pm$ 0.007} & \cellcolor{gray!10}\textbf{0.019 $\pm$ 0.005} & \cellcolor{gray!10}0.246 $\pm$ 0.297 & \cellcolor{yellow!30}\textbf{0.482 $\pm$ 0.003} & \cellcolor{gray!10}0.613 $\pm$ 0.210 & \cellcolor{yellow!30}\textbf{0.51 $\pm$ 0.07} & \cellcolor{gray!10}\textbf{0.73 $\pm$ 0.11} & \cellcolor{yellow!30}\textbf{0.88 $\pm$ 0.02} & \cellcolor{yellow!30}\textbf{0.94 $\pm$ 0.01} \\
    & Early AJEST & \cellcolor{yellow!30}\textbf{0.011 $\pm$ 0.003} & \cellcolor{yellow!30}\textbf{0.227 $\pm$ 0.008} & \cellcolor{yellow!30}\textbf{0.018 $\pm$ 0.004} & 0.110 $\pm$ 0.019 & \textbf{0.483 $\pm$ 0.003} & \textbf{0.516 $\pm$ 0.019} & \textbf{0.50 $\pm$ 0.08} & \cellcolor{yellow!30}\textbf{0.75 $\pm$ 0.08} & \cellcolor{yellow!30}\textbf{0.88 $\pm$ 0.02} & \textbf{0.93 $\pm$ 0.02} \\
    & \cellcolor{gray!10} Full AJEST & \cellcolor{gray!10}0.024 $\pm$ 0.006 & \cellcolor{gray!10}0.253 $\pm$ 0.009 & \cellcolor{gray!10}0.028 $\pm$ 0.008 & \cellcolor{gray!10}0.118 $\pm$ 0.025 & \cellcolor{yellow!30}\textbf{0.482 $\pm$ 0.004} & \cellcolor{gray!10}\textbf{0.514 $\pm$ 0.020} & \cellcolor{gray!10}0.40 $\pm$ 0.07 & \cellcolor{gray!10}0.53 $\pm$ 0.12 & \cellcolor{gray!10}0.80 $\pm$ 0.03 & \cellcolor{gray!10}0.85 $\pm$ 0.04 \\
    & Random & \textbf{0.013 $\pm$ 0.003} & \textbf{0.232 $\pm$ 0.005} & \textbf{0.019 $\pm$ 0.003} & \cellcolor{yellow!30}\textbf{0.101 $\pm$ 0.006} & \cellcolor{yellow!30}\textbf{0.482 $\pm$ 0.002} & \cellcolor{yellow!30}\textbf{0.508 $\pm$ 0.009} & \textbf{0.47 $\pm$ 0.06} & \textbf{0.71 $\pm$ 0.08} & \textbf{0.87 $\pm$ 0.02} & \textbf{0.93 $\pm$ 0.01} \\

    \midrule
    \multirow{4}{*}{\textbf{\begin{tabular}{c} Same \\ data \\ budget \end{tabular}}} 
    & \cellcolor{gray!10} Baseline & \cellcolor{gray!10}0.026 $\pm$ 0.008 & \cellcolor{gray!10}0.255 $\pm$ 0.011 & \cellcolor{gray!10}0.034 $\pm$ 0.010 & \cellcolor{gray!10}0.156 $\pm$ 0.053 & \cellcolor{gray!10}0.487 $\pm$ 0.003 & \cellcolor{gray!10}0.543 $\pm$ 0.036 & \cellcolor{gray!10}0.40 $\pm$ 0.07 & \cellcolor{gray!10}0.53 $\pm$ 0.15 & \cellcolor{gray!10}0.80 $\pm$ 0.05 & \cellcolor{gray!10}0.83 $\pm$ 0.09 \\
    & Early AJEST & 0.022 $\pm$ 0.007 & 0.249 $\pm$ 0.012 & 0.028 $\pm$ 0.006 & 0.120 $\pm$ 0.029 & \cellcolor{yellow!30}\textbf{0.481 $\pm$ 0.005} & \textbf{0.515 $\pm$ 0.030} & \textbf{0.42 $\pm$ 0.07} & 0.58 $\pm$ 0.14 & 0.81 $\pm$ 0.05 & 0.86 $\pm$ 0.06 \\
    & \cellcolor{gray!10} Full AJEST & \cellcolor{gray!10}\textbf{0.013 $\pm$ 0.003} & \cellcolor{gray!10}\textbf{0.234 $\pm$ 0.008} & \cellcolor{yellow!30}\textbf{0.018 $\pm$ 0.004} & \cellcolor{gray!10}\textbf{0.102 $\pm$ 0.006} & \cellcolor{yellow!30}\textbf{0.481 $\pm$ 0.002} & \cellcolor{yellow!30}\textbf{0.507 $\pm$ 0.009} & \cellcolor{gray!10}\textbf{0.46 $\pm$ 0.06} & \cellcolor{gray!10}\textbf{0.69 $\pm$ 0.10} & \cellcolor{gray!10}\textbf{0.86 $\pm$ 0.03} & \cellcolor{gray!10}\textbf{0.92 $\pm$ 0.02} \\
    & Random & \cellcolor{yellow!30}\textbf{0.012 $\pm$ 0.002} & \cellcolor{yellow!30}\textbf{0.231 $\pm$ 0.005} & \textbf{0.019 $\pm$ 0.003} & \cellcolor{yellow!30}\textbf{0.101 $\pm$ 0.005} & \textbf{0.482 $\pm$ 0.001} & \textbf{0.508 $\pm$ 0.005} & \cellcolor{yellow!30}\textbf{0.47 $\pm$ 0.06} & \cellcolor{yellow!30}\textbf{0.72 $\pm$ 0.07} & \cellcolor{yellow!30}\textbf{0.87 $\pm$ 0.02} & \cellcolor{yellow!30}\textbf{0.93 $\pm$ 0.01} \\
    
    \bottomrule
  \end{tabular}%
  }
\end{table*}

\noindent

\noindent

\subsection{EDM2 for Tiny ImageNet}
\label{sec:supp-results-images}

\begin{figure*}
  \centerline{\includegraphics[width=0.72\textwidth]{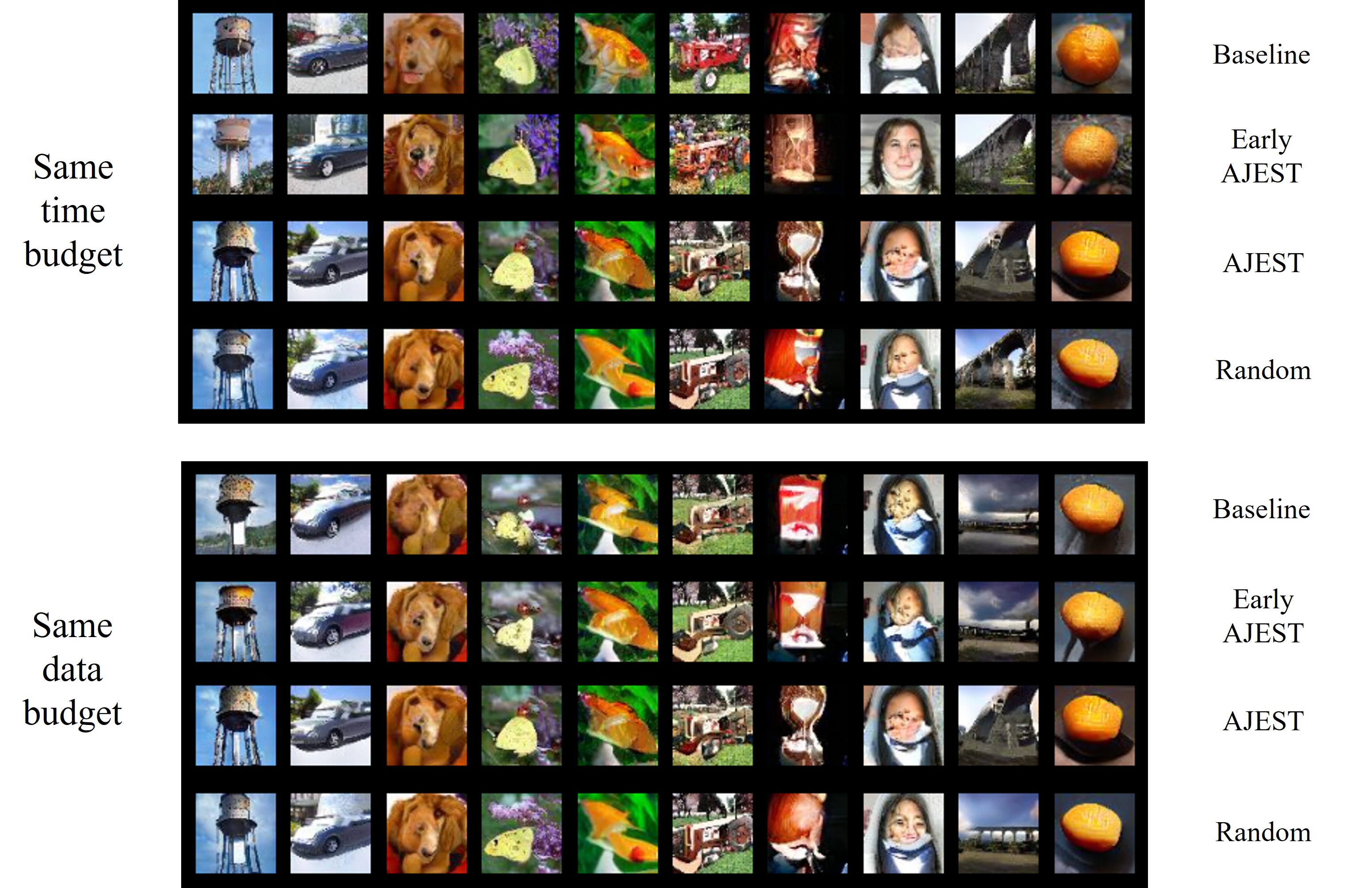}}
  \caption{Generated images for 10 classes on the 48 hs time-limited and 7.1 Mimg data-limited scenarios: (1) water tower, (2) convertible, (3) golden retriever, (4) sulphur butterfly, (5) goldfish, (6) tractor, (7) hourglass, (8) neck brace, (9) viaduct, (10) orange.}
  \label{fig:generated-images}
\end{figure*}

We present images generated on the fix-budget and fix-time scenarios in Figure \ref{fig:generated-images}. In the same-time comparison, it is remarkable for Early AJEST to lead to the first human face and the first goldfish with eyes. The autoguidance baseline, however, leads to the first tractor and the closest image to a golden retriever's face. 

\noindent

\begin{figure*}
  \centerline{\includegraphics[width=0.72\textwidth]{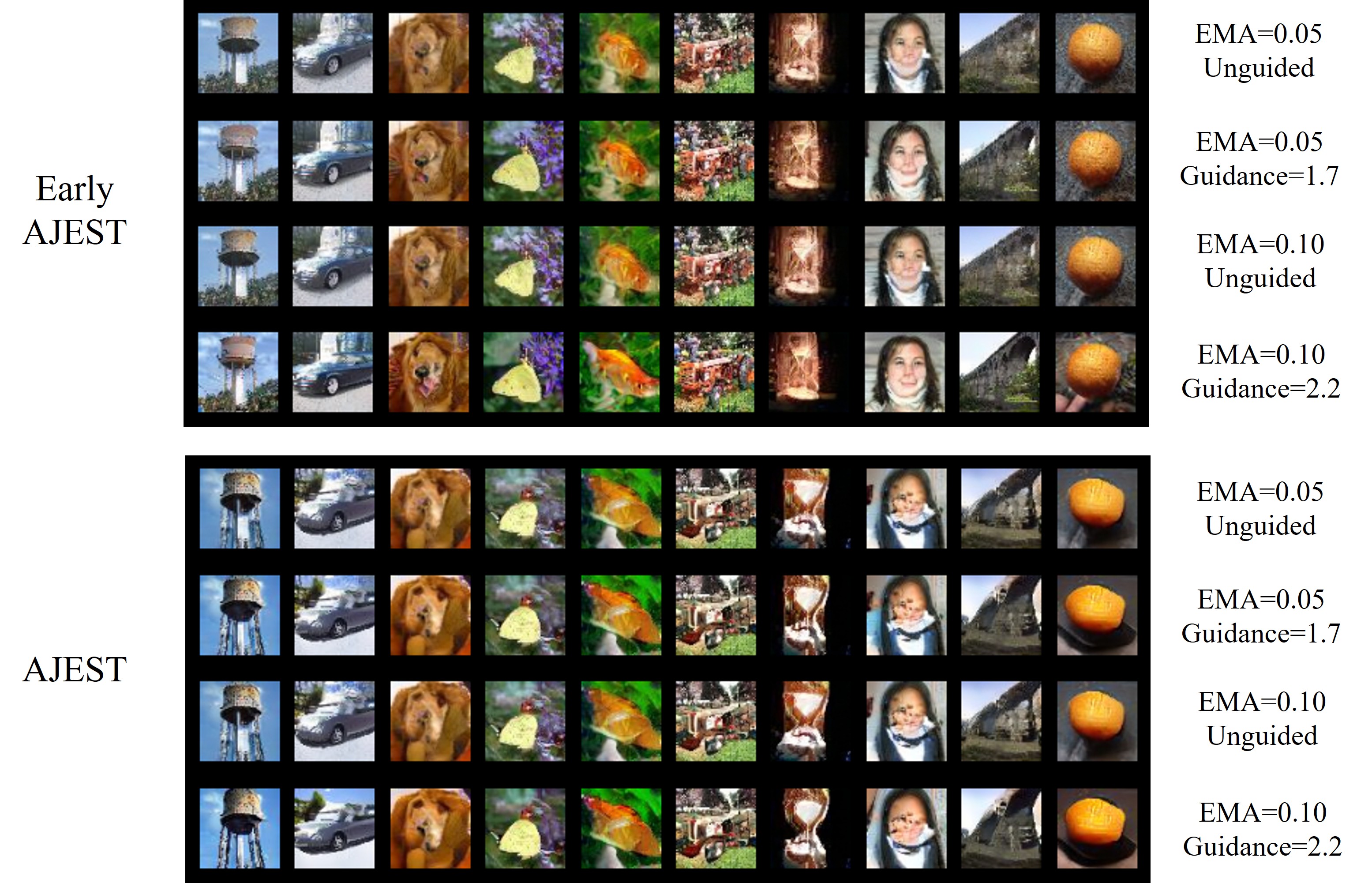}}
  \caption{Generated images for 10 classes for AJEST under two different EMA values, both with guidance and without it: (1) water tower, (2) convertible, (3) golden retriever, (4) sulphur butterfly, (5) goldfish, (6) tractor, (7) hourglass, (8) neck brace, (9) viaduct, (10) orange.}
  \label{fig:generated-images-hyper}
\end{figure*}

We also present images generated using different EMA and guidance weight in Figure \ref{fig:generated-images-hyper}. These hyperparameters are shown to have a large influence on the image quality, as first reported by Karras et al. \cite{2024KarrasAutoGuidance}. Guidance improves sharpness and contents considerably for EMA=0.10, but not as much for EMA=0.05. This might be due to 2.2 being closer to the ideal EMA for 0.10 than 1.7 is for 0.05.

\noindent

Tables with separate results for EMA=0.05 and EMA=0.10 -both with guidance and without it- can be seen in Tables \ref{tab:images_0.10} and \ref{tab:images_0.10}. Results averaged over those two EMA, but with both guided and unguided metrics, are presented in Table \ref{tab:images_mean_full}.

Our Tiny ImageNet results show new evidence that autoguidance is especially beneficial to improve coverage and precision over the ground-truth distribution: guidance improves all Swin-L top-1 and top-5 results, for as much as 10\% in the best-case scenario.

In the time-limited scenario, there is a marked difference between FID and all other metrics. Random reaches the best FID score, and Early AJEST outperforms autoguidance. However, all other metrics show Baseline as the best, followed by Early AJEST, Random and finally AJEST. This may be added to the list of alarming evidence on how FID might not be an adequate metric to measure generated image quality \cite{2023SteinFD}. We strongly believe that generative vision AI would benefit from better metric definitions.

\begin{table*}[h!]
  \tiny
  \centering
  \caption{Evaluation metrics on Tiny ImageNet with EMA=0.05. Yellow-filled cells indicate the best scores for each block of the table, considering both unguided results and guided results using guidance weight 1.7. Bold values indicate the best results for guided and unguided models separately.}
  \label{tab:images_0.05}
  \resizebox{\textwidth}{!}{%
  \begin{tabular}{ll
                  cc
                  cc
                  cc
                  cc}
    \toprule
    \multirow{3}{*}{\textbf{Comparison}} & \multirow{3}{*}{\textbf{Method}} 
    & \multicolumn{4}{c}{\textbf{Perceptual metrics}} 
    & \multicolumn{4}{c}{\textbf{Classification-based metrics}} \\
    
    \cmidrule(lr){3-6} \cmidrule(lr){7-10}
    & & \multicolumn{2}{c}{\textbf{FID}} 
      & \multicolumn{2}{c}{\textbf{FD-DINOv2}} 
      & \multicolumn{2}{c}{\textbf{Top-1}} 
      & \multicolumn{2}{c}{\textbf{Top-5}} \\
    
    \cmidrule(lr){3-4} \cmidrule(lr){5-6} \cmidrule(lr){7-8} \cmidrule(lr){9-10}
    & & \textbf{Unguided} & \textbf{Guided} 
      & \textbf{Unguided} & \textbf{Guided} 
      & \textbf{Unguided} & \textbf{Guided} 
      & \textbf{Unguided} & \textbf{Guided} \\
    
    \midrule
    \multirow{4}{*}{\textbf{\begin{tabular}{c} Same \\ time \\ budget \end{tabular}}} 
    & \cellcolor{gray!10} Baseline     & \cellcolor{gray!10}44.6 & \cellcolor{gray!10}34.3 & \cellcolor{gray!10}822 & \cellcolor{yellow!30}\textbf{677} & \cellcolor{gray!10}\textbf{47.9} & \cellcolor{yellow!30}\textbf{56.8} & \cellcolor{gray!10}\textbf{69.8} & \cellcolor{gray!10}78.5 \\
    & Early AJEST  & 45.5 & 34.7 & \textbf{813} & \cellcolor{yellow!30}\textbf{677} & 46.3 & 54.2 & 68.9 & \cellcolor{yellow!30}\textbf{76.8} \\
    & \cellcolor{gray!10} AJEST        & \cellcolor{gray!10}42.7 & \cellcolor{gray!10}41.8 & \cellcolor{gray!10}959 & \cellcolor{gray!10}959 & \cellcolor{gray!10}29.3 & \cellcolor{gray!10}31.2 & \cellcolor{gray!10}54.2 & \cellcolor{gray!10}53.7 \\
    & Random       & \textbf{33.8} & \cellcolor{yellow!30}\textbf{28.6} & 799 & 717 & 39.2 & 44.9 & 62.9 & 67.6 \\
    
    \midrule
    \multirow{4}{*}{\textbf{\begin{tabular}{c} Same \\ data \\ budget \end{tabular}}} 
    & \cellcolor{gray!10} Baseline     & \cellcolor{gray!10}41.5 & \cellcolor{gray!10}40.8 & \cellcolor{gray!10}927 & \cellcolor{gray!10}927 & \cellcolor{gray!10}\textbf{47.7} & \cellcolor{yellow!30}\textbf{58.4} & \cellcolor{gray!10}\textbf{70.3} & \cellcolor{gray!10}77.0 \\
    & Early AJEST  & 38.8 & 36.1 & 878 & 857 & 45.3 & 57.1 & 69.6 & \cellcolor{yellow!30}\textbf{78.3} \\
    & \cellcolor{gray!10} AJEST        & \cellcolor{gray!10}42.7 & \cellcolor{gray!10}41.8 & \cellcolor{gray!10}959 & \cellcolor{gray!10}959 & \cellcolor{gray!10}29.3 & \cellcolor{gray!10}31.2 & \cellcolor{gray!10}54.2 & \cellcolor{gray!10}53.7 \\
    & Random       & \textbf{34.9} & \cellcolor{yellow!30}\textbf{30.4} & \textbf{823} & \cellcolor{yellow!30}\textbf{753} & 35.9 & 41.0 & 60.5 & 64.6 \\

    \bottomrule
  \end{tabular}%
  }
\end{table*}

\begin{table*}[h!]
  \tiny
  \centering
  \caption{Evaluation metrics on Tiny ImageNet with EMA=0.10. Yellow-filled cells indicate the best scores for each block of the table, considering both unguided results and guided results using guidance weight 2.2. Bold values indicate the best results for guided and unguided models separately.}
  \label{tab:images_0.10}
  \resizebox{\textwidth}{!}{%
  \begin{tabular}{ll
                  cc
                  cc
                  cc
                  cc}
    \toprule
    \multirow{3}{*}{\textbf{Comparison}} & \multirow{3}{*}{\textbf{Method}} 
    & \multicolumn{4}{c}{\textbf{Perceptual metrics}} 
    & \multicolumn{4}{c}{\textbf{Classification-based metrics}} \\
    
    \cmidrule(lr){3-6} \cmidrule(lr){7-10}
    & & \multicolumn{2}{c}{\textbf{FID}} 
      & \multicolumn{2}{c}{\textbf{FD-DINOv2}} 
      & \multicolumn{2}{c}{\textbf{Top-1}} 
      & \multicolumn{2}{c}{\textbf{Top-5}} \\
    
    \cmidrule(lr){3-4} \cmidrule(lr){5-6} \cmidrule(lr){7-8} \cmidrule(lr){9-10}
    & & \textbf{Unguided} & \textbf{Guided} 
      & \textbf{Unguided} & \textbf{Guided} 
      & \textbf{Unguided} & \textbf{Guided} 
      & \textbf{Unguided} & \textbf{Guided} \\
    
    \midrule
    \multirow{4}{*}{\textbf{\begin{tabular}{c} Same \\ time \\ budget \end{tabular}}} 
    & \cellcolor{gray!10} Baseline     & \cellcolor{gray!10}40.4 & \cellcolor{gray!10}26.9 & \cellcolor{gray!10}774 & \cellcolor{yellow!30}\textbf{571} & \cellcolor{gray!10}\textbf{51.9} & \cellcolor{gray!10}62.8 & \cellcolor{gray!10}\textbf{72.5} & \cellcolor{gray!10}82.7 \\
    & Early AJEST  & 40.9 & 27.2 & \textbf{769} & 578 & 48.2 & \cellcolor{yellow!30}\textbf{63.4} & 72.2 & \cellcolor{yellow!30}\textbf{82.9} \\
    & \cellcolor{gray!10} AJEST        & \cellcolor{gray!10}41.5 & \cellcolor{gray!10}40.3 & \cellcolor{gray!10}950 & \cellcolor{gray!10}940 & \cellcolor{gray!10}29.4 & \cellcolor{gray!10}32.5 & \cellcolor{gray!10}55.9 & \cellcolor{gray!10}55.5 \\
    & Random       & \textbf{32.8} & \cellcolor{yellow!30}\textbf{26.4} & 790 & 681 & 40.7 & 45.9 & 63.3 & 69.9 \\
    
    \midrule
    \multirow{4}{*}{\textbf{\begin{tabular}{c} Same \\ data \\ budget \end{tabular}}} 
    & \cellcolor{gray!10} Baseline     & \cellcolor{gray!10}41.1 & \cellcolor{gray!10}40.1 & \cellcolor{gray!10}930 & \cellcolor{gray!10}940 & \cellcolor{gray!10}\textbf{50.4} & \cellcolor{yellow!30}\textbf{65.3} & \cellcolor{gray!10}\textbf{72.7} & \cellcolor{gray!10}82.2 \\
    & Early AJEST  & 37.7 & 33.4 & 871 & 842 & 47.7 & 64.3 & 71.9 & \cellcolor{yellow!30}\textbf{82.7} \\
    & \cellcolor{gray!10} AJEST        & \cellcolor{gray!10}41.5 & \cellcolor{gray!10}40.3 & \cellcolor{gray!10}950 & \cellcolor{gray!10}940 & \cellcolor{gray!10}29.4 & \cellcolor{gray!10}32.5 & \cellcolor{gray!10}55.9 & \cellcolor{gray!10}55.5 \\
    & Random       & \textbf{34.0} & \cellcolor{yellow!30}\textbf{28.1} & \textbf{815} & \cellcolor{yellow!30}\textbf{720} & 36.0 & 40.8 & 60.7 & 65.4 \\

    \bottomrule
  \end{tabular}%
  }
\end{table*}

\begin{table*}[h!]
  \tiny
  \centering
  \caption{Evaluation metrics on Tiny ImageNet, averaging results from EMA=0.05 and EMA=0.10. Bold yellow values indicate the best scores for each block of the table, considering both unguided results and guided results using guidance weight 1.7 for EMA=0.05 and 2.2 for EMA=0.10. Bold values indicate the best results for guided and unguided models separately.}
  \label{tab:images_mean_full}
  \resizebox{\textwidth}{!}{%
  \begin{tabular}{ll
                  cc
                  cc
                  cc
                  cc}
    \toprule
    \multirow{3}{*}{\textbf{Comparison}} & \multirow{3}{*}{\textbf{Method}} 
    & \multicolumn{4}{c}{\textbf{Perceptual metrics}} 
    & \multicolumn{4}{c}{\textbf{Classification-based metrics}} \\
    
    \cmidrule(lr){3-6} \cmidrule(lr){7-10}
    & & \multicolumn{2}{c}{\textbf{FID}} 
      & \multicolumn{2}{c}{\textbf{FD-DINOv2}} 
      & \multicolumn{2}{c}{\textbf{Top-1}} 
      & \multicolumn{2}{c}{\textbf{Top-5}} \\
    
    \cmidrule(lr){3-4} \cmidrule(lr){5-6} \cmidrule(lr){7-8} \cmidrule(lr){9-10}
    & & \textbf{Unguided} & \textbf{Guided} 
      & \textbf{Unguided} & \textbf{Guided} 
      & \textbf{Unguided} & \textbf{Guided} 
      & \textbf{Unguided} & \textbf{Guided} \\
    
    \midrule
    \multirow{4}{*}{\textbf{Same time budget}} 
    & \cellcolor{gray!10} Baseline     & \cellcolor{gray!10}42.5 & \cellcolor{gray!10}30.6 & \cellcolor{gray!10}798 & \cellcolor{yellow!30}\textbf{624} & \cellcolor{gray!10}\textbf{\textcolor{black}{49.9}} & \cellcolor{yellow!30}\textbf{59.8} & \cellcolor{gray!10}\textbf{71.1} & \cellcolor{yellow!30}\textbf{80.6} \\
    & Early AJEST  & 43.2 & \textbf{31.0} & \textbf{791} & \textbf{628} & 47.3 & \textbf{58.8} & 70.6 & 79.8 \\
    & \cellcolor{gray!10} AJEST        & \cellcolor{gray!10}42.1 & \cellcolor{gray!10}41.0 & \cellcolor{gray!10}955 & \cellcolor{gray!10}949 & \cellcolor{gray!10}29.3 & \cellcolor{gray!10}31.8 & \cellcolor{gray!10}55.0 & \cellcolor{gray!10}54.6 \\
    & Random       & \textbf{33.3} & \cellcolor{yellow!30}\textbf{27.5} & 795 & 699 & 39.9 & 45.4 & 63.1 & 68.8 \\
    
    \midrule
    \multirow{4}{*}{\textbf{Same data budget}} 
    & \cellcolor{gray!10} Baseline     & \cellcolor{gray!10}41.3 & \cellcolor{gray!10}40.4 & \cellcolor{gray!10}929 & \cellcolor{gray!10}934 & \cellcolor{gray!10}\textbf{49.0} & \cellcolor{yellow!30}\textbf{61.8} & \cellcolor{gray!10}\textbf{71.5} & \cellcolor{gray!10}\textbf{79.6} \\
    & Early AJEST  & 38.2 & \textbf{34.7} & 875 & 849 & 46.5 & \textbf{60.7} & 70.8 & \cellcolor{yellow!30}\textbf{80.5} \\
    & \cellcolor{gray!10} AJEST        & \cellcolor{gray!10}42.1 & \cellcolor{gray!10}41.0 & \cellcolor{gray!10}955 & \cellcolor{gray!10}949 & \cellcolor{gray!10}29.3 & \cellcolor{gray!10}31.8 & \cellcolor{gray!10}55.0 & \cellcolor{gray!10}54.6 \\
    & Random       & \textbf{{34.4}} & \cellcolor{yellow!30}\textbf{29.2} & \textbf{819} & \cellcolor{yellow!30}\textbf{737} & 36.0 & 40.9 & 60.6 & 65.0 \\

    \bottomrule
  \end{tabular}%
  }
\end{table*}

\begin{figure}[hb]
  \centerline{\includegraphics[width=0.5\textwidth]{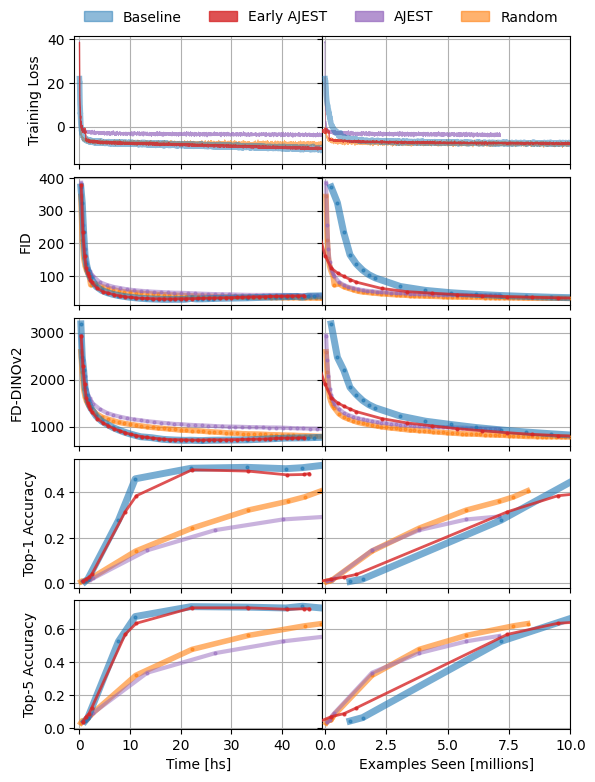}}
  \caption{Training loss and validation metrics on Tiny ImageNet for EMA=0.10 with no guidance.}
  \label{fig:images-0.1}
\end{figure}

\begin{figure}[hb]
  \centerline{\includegraphics[width=0.5\textwidth]{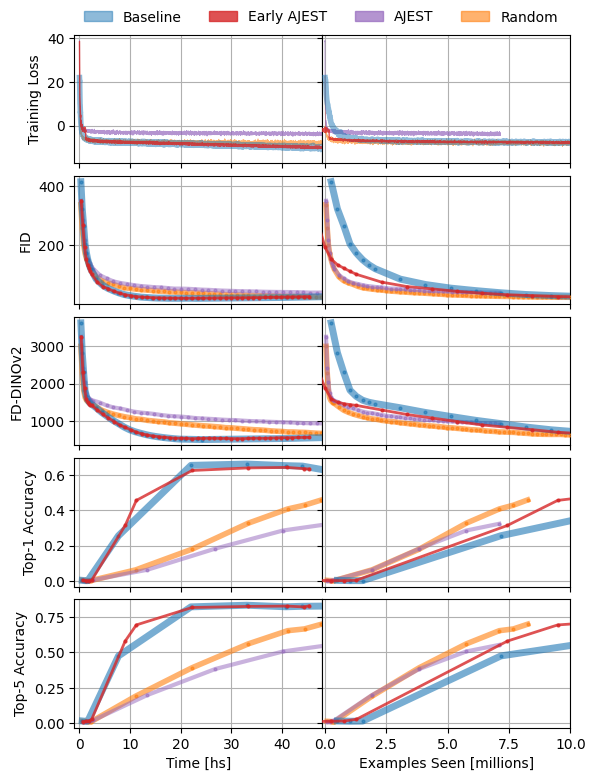}}
  \caption{Training loss and validation metrics on Tiny ImageNet for EMA=0.10 with guidance weight $\alpha=2.2$.}
  \label{fig:images-0.1-guided}
\end{figure}

\begin{figure}[hb]
  \centerline{\includegraphics[width=0.5\textwidth]{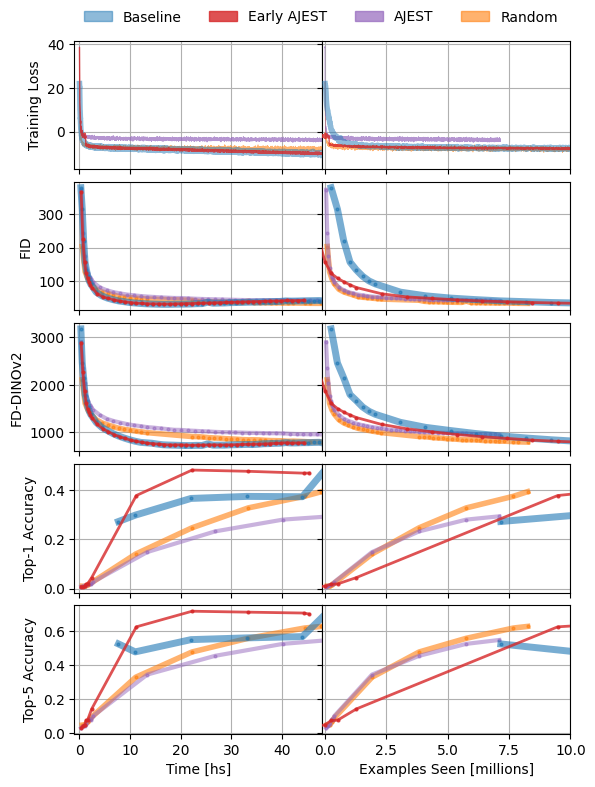}}
  \caption{Training loss and validation metrics on Tiny ImageNet for EMA=0.05 with no guidance.}
  \label{fig:images-0.05}
\end{figure}

\begin{figure}[hb]
  \centerline{\includegraphics[width=0.5\textwidth]{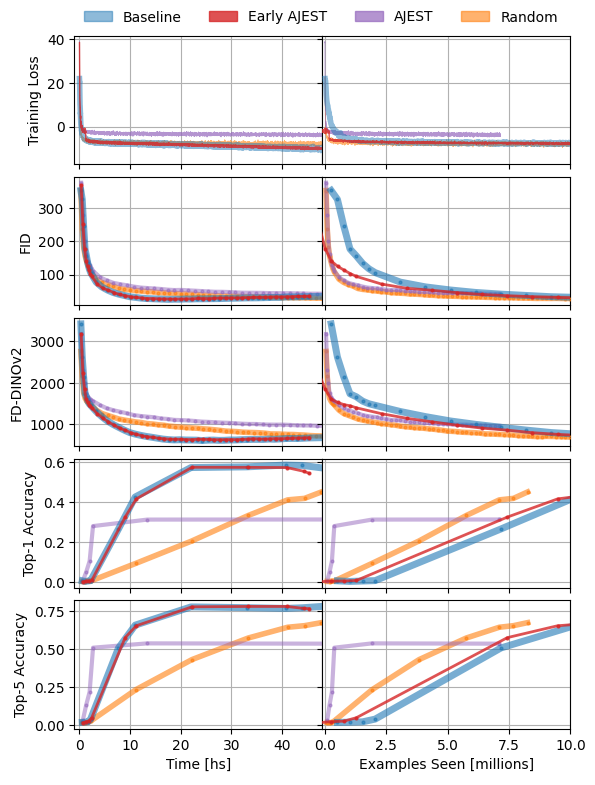}}
  \caption{Training loss and validation metrics on Tiny ImageNet for EMA=0.05 and guidance weight $\alpha=1.7$.}
  \label{fig:images-0.05-guided}
\end{figure}

We include training and validation curves for different EMA and guidance combinations in Figures \ref{fig:images-0.1} to \ref{fig:images-0.05-guided}. The trigger for Early AJEST can be deduced from the training curves, as its training loss falls from AJEST's to Baseline's. Data selection methods lead to consistently better results at low data budgets, especially Full AJEST and random data selection. Early AJEST only shows limited-data advantages over Baseline on FID and FD-DINOv2, but it follows quite closely its Top-1 and Top-5 performance under limited-time settings. 

Evidence suggests that data selection might help mitigate overfitting on FID and FD-DINOv2 for unguided models: this is especially noticeable on the FID plot for EMA=0.05 and $\alpha=1.7$ (Figure \ref{fig:images-0.05-guided}). However, this effect might also be explained by this being a very early stage in training for AJEST and Random; the same overfitting phenomenum might be present at later training times.

\noindent

\noindent

\noindent

\noindent

\subsection{2D toy example and other AJEST strategies}

We explored the possibility of using both JEST's learnability score and an inverted learnability score:

\vspace{-3mm}\begin{equation}
    \label{eq:inverted}
    s_{ij}^{inv}=L^{R}_{j}-L_{i}^{L}=-s_{ij}^{learn}
\end{equation}

In JEST they use \cref{eq:learnability} to prioritize extremely difficult samples that are only easy to learn for a high-capacity reference \cite{2024EvansJEST}. This learnability score is therefore designed for situations in which a larger pre-trained model with good performance is available to use it as the smaller learner model's teacher. Karras et al's research was framed on a different situation, one in which we do not count with a different, better pre-trained model, but with a "bad version" of the learner model instead: a copy of the learner model with restricted capability \cite{2024KarrasAutoGuidance}. We therefore hypothesized that it may be necessary to invert the sign of the learnability score, using \cref{eq:inverted} to compensate for the guide not being more capable than the learner. We refer to this approach as inverted JEST or iJEST.

\begin{figure}[h]
  \centerline{\includegraphics[width=0.35\textwidth]{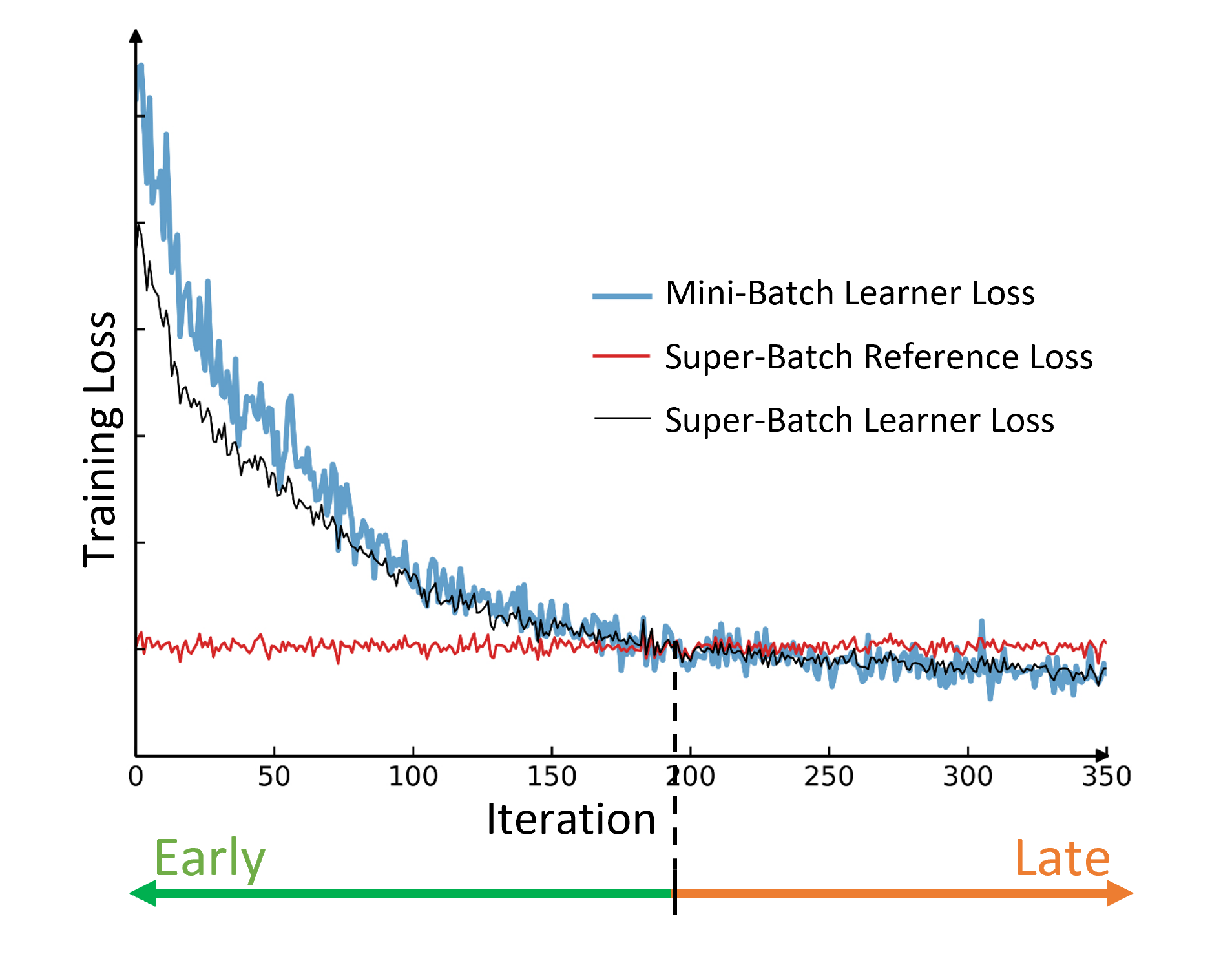}}
  \caption{Illustration of how Early and Late data selection strategies are applied during training. Early in training, the reference is better than the learner, so the super-batch reference loss is smaller than the super-batch learner loss. Once the learner becomes better than the reference, the situation is inverted.}
  \label{fig:early-late}
\end{figure}

In addition to Early AJEST, we also explored a Late AJEST strategy. As indicated in Figure \ref{fig:early-late}, data selection would only be run late in training under the Late AJEST strategy. We assumed that Early AJEST and Late iAJEST may reach complementary results, especially when combined with the inverted and non-inverted learnability scores. If data selection with learnability scores required a large pre-trained model to work correctly, then we would expect to see it work better early in training (Early AJEST). In contrast, in case the inverted learnability scores would help alleviate this JEST requirement, then we would expect iAJEST to work better late in training (Late iAJEST).

\noindent

Inverted JEST learnability scores applied late in training can lead to improvements superior to 5\% both for the average L2 distance and the mandala score. In general, iJEST seems to improve the coverage over the generated samples distribution, meaning it can lead to generative models with higher diversity. Despite this, applying iJEST late in training for an extended period of time seems to over-estimate the autoguidance corrective force, resulting in lower average loss and L2 distance evaluated in the full tree. This seems to indicate that iJEST scoring enforces a trade-off between diversity and fidelity or precision. The same trade-off is apparently not as strongly enforced by non-inverted JEST scores, allowing Early JEST to improve autoguidance metrics without any marked disadvantages.

The best results were obtained applying Early AJEST. The cost of applying JEST in this way was an increase of just 2.7\% in the training time, because the data selection strategy was only applied to 190 of
the 4096 training iterations. Early JEST improvements were less marked whenever guidance interpolation was applied in the inference phase, indicating that JEST may be able to use the small less-capable reference
model only during training, discarding it at the inference phase.

Despite the high filtering ratio of 80\%, our Early JEST method does not greatly reduce the volume of data used for training. Applied for 194 iterations on a super-batch size B of 4096, Early JEST implies only 1\% of the total number of data points are discarded. In
consequence, it may seem that Early JEST is not particularly data-efficient. Its success does not contradict the notion that diffusion models cannot learn a distribution accurately enough if a significant number of examples are discarded. However, the success of Early JEST in the 2D
tree task does seem to imply that there are better ways to use data for diffusion models to learn. A slightly more data-efficient algorithm may still be more data-efficient that one that samples data uniformly from the available distribution.

\begin{table*}[h!]
  \tiny
  \centering
  \caption{Evaluation metrics for models trained for 4096 iterations on the 2D tree task. Results are averaged over 5 runs with different random seeds, with standard deviation indicating uncertainty. Bold yellow values represent the best scores for each metric across guided and unguided settings. Bold black values highlight scores statistically indistinguishable from the best (within uncertainty).}
  \label{tab:toy_full_reformatted}
  \resizebox{\textwidth}{!}{%
  \begin{tabular}{cc
                  cc
                  cc
                  cc
                  cc
                  cc}
    \toprule
    \multirow{3}{*}{\textbf{Comparison}} & \multirow{3}{*}{\textbf{Method}} 
    & \multicolumn{2}{c}{\textbf{Average Loss (Unguided)}} 
    & \multicolumn{2}{c}{\textbf{L2 Distance (All)}} 
    & \multicolumn{2}{c}{\textbf{L2 Distance (External)}} 
    & \multicolumn{2}{c}{\textbf{Mandala Score}} 
    & \multicolumn{2}{c}{\textbf{Classification Score}} \\

    \cmidrule(lr){3-4} \cmidrule(lr){5-6} \cmidrule(lr){7-8} \cmidrule(lr){9-10} \cmidrule(lr){11-12}
    & & \textbf{All Branches} & \textbf{External Branches} 
      & \textbf{Unguided} & \textbf{Guided} 
      & \textbf{Unguided} & \textbf{Guided} 
      & \textbf{Unguided} & \textbf{Guided} 
      & \textbf{Unguided} & \textbf{Guided} \\

    \midrule
    \multirow{1}{*}{\textbf{Autoguidance}} 
    & \textbf{Autoguidance} & \cellcolor{yellow!30}\textbf{0.0111 $\pm$0.003} & \textbf{0.223 $\pm$0.007} & 0.0180 $\pm$0.004 & 0.207 $\pm$0.220 & 0.4827 $\pm$0.003 & 0.603 $\pm$0.187 & \cellcolor{yellow!30}\textbf{0.63 $\pm$0.05} & \cellcolor{yellow!30}\textbf{0.82 $\pm$0.06} & \cellcolor{yellow!30}\textbf{0.89 $\pm$0.02} & \cellcolor{yellow!30}\textbf{0.94 $\pm$0.01} \\

    \midrule
    \multirow{2}{*}{\textbf{Early}} 
    & AJEST   & 0.0108 $\pm$0.003 & \cellcolor{yellow!30}\textbf{0.222 $\pm$0.007} & 0.0170 $\pm$0.003 & 0.106 $\pm$0.014 & 0.4830 $\pm$0.004 & 0.516 $\pm$0.019 & \cellcolor{yellow!30}\textbf{0.63 $\pm$0.05} & \textbf{0.81 $\pm$0.04} & \cellcolor{yellow!30}\textbf{0.89 $\pm$0.02} & \textbf{0.93 $\pm$0.01} \\
    & iAJEST  & \textbf{0.0112 $\pm$0.003} & \textbf{0.223 $\pm$0.007} & \cellcolor{yellow!30}\textbf{0.0162 $\pm$0.002} & 0.108 $\pm$0.020 & 0.4823 $\pm$0.004 & 0.521 $\pm$0.031 & \cellcolor{yellow!30}\textbf{0.63 $\pm$0.05} & \cellcolor{yellow!30}\textbf{0.82 $\pm$0.03} & \textbf{0.88 $\pm$0.02} & \textbf{0.92 $\pm$0.01} \\

    \midrule
    \multirow{2}{*}{\textbf{Late}} 
    & AJEST   & 0.0138 $\pm$0.003 & \textbf{0.229 $\pm$0.007} & 0.0176 $\pm$0.003 & \cellcolor{yellow!30}\textbf{0.100 $\pm$0.004} & 0.4819 $\pm$0.002 & \textbf{0.510 $\pm$0.009} & \textbf{0.57 $\pm$0.05} & 0.76 $\pm$0.06 & 0.86 $\pm$0.02 & 0.92 $\pm$0.01 \\
    & iAJEST  & 0.0129 $\pm$0.003 & \textbf{0.228 $\pm$0.007} & 0.0196 $\pm$0.004 & 0.114 $\pm$0.024 & 0.4819 $\pm$0.002 & 0.522 $\pm$0.027 & \textbf{0.62 $\pm$0.06} & 0.78 $\pm$0.06 & \textbf{0.86 $\pm$0.02} & \textbf{0.93 $\pm$0.01} \\

    \midrule
    \multirow{2}{*}{\textbf{Full}} 
    & AJEST   & 0.0138 $\pm$0.003 & 0.231 $\pm$0.010 & 0.0181 $\pm$0.004 & \textbf{0.101 $\pm$0.004} & 0.4814 $\pm$0.002 & \textbf{0.508 $\pm$0.009} & \textbf{0.57 $\pm$0.06} & 0.74 $\pm$0.06 & \textbf{0.86 $\pm$0.02} & 0.92 $\pm$0.01 \\
    & iAJEST  & 0.0131 $\pm$0.004 & \textbf{0.229 $\pm$0.010} & 0.0186 $\pm$0.004 & \textbf{0.101 $\pm$0.004} & \cellcolor{yellow!30}\textbf{0.4801 $\pm$0.003} & \cellcolor{yellow!30}\textbf{0.502 $\pm$0.013} & \textbf{0.61 $\pm$0.06} & 0.78 $\pm$0.06 & \textbf{0.88 $\pm$0.02} & 0.92 $\pm$0.01 \\

    \midrule
    \multirow{1}{*}{\textbf{Random}} 
    & Random  & 0.0124 $\pm$0.002 & 0.231 $\pm$0.005 & 0.0192 $\pm$0.003 & \textbf{0.101 $\pm$0.005} & 0.4819 $\pm$0.001 & \textbf{0.508 $\pm$0.005} & 0.47 $\pm$0.06 & 0.72 $\pm$0.07 & \textbf{0.87 $\pm$0.02} & \textbf{0.93 $\pm$0.01} \\

    \bottomrule
  \end{tabular}%
  }
\end{table*}

\section{Metrics definition}
\label{sec:metrics}

To quantitatively evaluate the performance of the diffusion models and the impact of data selection strategies, we employ several complementary metrics to capture both fidelity to the ground truth and diversity of the generated samples.

\subsection{Metrics for the 2D tree toy example}

We used 163,840 generated points with the same random seeds to evaluate 2D tree metrics.

\paragraph{Average Loss.}
The primary training and evaluation loss is the mean squared error (MSE) between the model's predicted score function and the ground truth score, scaled by the noise level $\sigma^2$:

\begin{equation}
\mathcal{L}_{\mathrm{MSE}} = \frac{1}{N} \sum_{i=1}^N \sigma_i^2 \left\| S(\mathbf{x}_i, \sigma_i) - s^{gt}(\mathbf{x}_i, \sigma_i) \right\|^2
\label{eq:loss}
\end{equation}
where $N$ is the batch size, $\mathbf{x}_i$ is a data sample, $\sigma_i$ is the noise level, $S$ is the predicted score with or without guidance, and $s^{gt}$ is the ground truth score.

This loss is computed both over the entire data distribution and specifically on the external branches of the 2D tree, which correspond to regions of lower data density and higher generation difficulty. Lower loss values indicate better alignment with the true data distribution.

\paragraph{L2 Distance.}
To assess the quality of generated samples, we compute the average Euclidean (L2) distance between fully denoised samples produced by the model, $\mathbf{x}_j$, and samples from the ground truth distribution, $\mathbf{x}_j^{\mathrm{gt}}$ obtained with the same random seed. This metric is reported both for the full distribution and for the external branches, providing insight into the model's ability to capture both the core and the periphery of the data manifold. We also report a "guided" L2 distance, where the model's outputs are generated with guidance (e.g., autoguidance or classifier-free guidance) during sampling. The L2 distance is given by:
\vspace{-3mm}\begin{equation}
\mathrm{L2\ Distance} = \frac{1}{M} \sum_{j=1}^M \left\| \mathbf{x}_j - \mathbf{x}_j^{\mathrm{gt}} \right\|
\end{equation}

\vspace{-3mm}where $M$ is the number of generated samples, $\mathbf{x}_j$ is a generated (denoised) sample, and $\mathbf{x}_j^{\mathrm{gt}}$ is the corresponding ground truth sample.

\paragraph{Mandala Score.}
To measure the diversity and coverage of the generated samples, we introduce the "mandala score". The 2D data space is discretized into a grid of $K$ cells. Let $\mathcal{C}_{\mathrm{gt}}$ denote the set of grid cells covered by the ground truth distribution, and let $\mathcal{C}_{\mathrm{gen}}$ denote the set of $100\times100$ grid cells that contain at least one generated sample. The mandala score is then defined as
\vspace{-3mm}\begin{equation}
\mathrm{Mandala\ Score} = \frac{|\mathcal{C}_{\mathrm{gt}} \cap \mathcal{C}_{\mathrm{gen}}|}{|\mathcal{C}_{\mathrm{gt}}|}
\end{equation}

\vspace{-3mm}where $|\cdot|$ denotes the cardinality of a set. A higher mandala score indicates better coverage and diversity, while a lower score suggests mode collapse or insufficient exploration of the data space.

\noindent\begin{figure}
  \centerline{\includegraphics[width=0.5\textwidth]{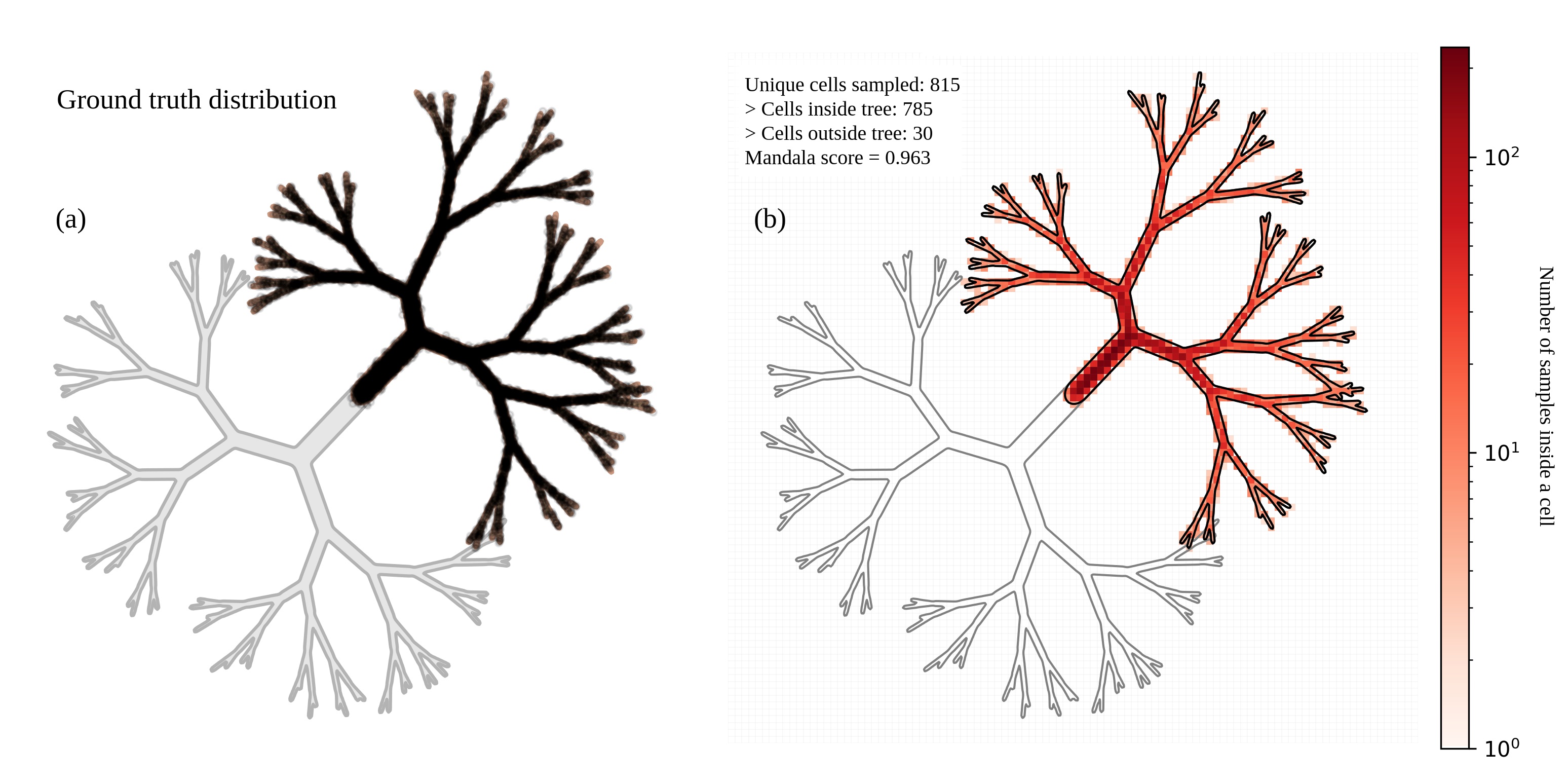}}
  \caption{Single-run mandala score calculation for the 2D tree ground truth data distribution: (a) $2^{14}=16384$ data points are generated as plotted on the left; (b) a section of the 2D space that completely covers the data distribution is discretized into a grid with $K=100^2$ cells; the number of data points belonging to each cell is counted and plotted on the right.}
  \label{fig:mandala-GT}
\end{figure}

\paragraph{Classification Metric.}
We also report a classification-based metric, specifically the binary accuracy. Let $y_j$ be the true class label for sample $j$ and $\hat{y}_j$ the predicted class label. The binary accuracy is defined as:
\vspace{-3mm}\begin{equation}
\mathrm{Accuracy} = \frac{1}{M} \sum_{j=1}^M \mathbb{I}(\hat{y}_j = y_j)
\end{equation}

\vspace{-3mm}where $\mathbb{I}(\cdot)$ is the indicator function and $M$ is the number of samples. Higher accuracy indicates that the model generates samples that are both realistic and class-consistent.

\subsection{Metrics for image generation in Tiny ImageNet}

\paragraph{Quality evaluation.}

Following Karras et al. \cite{2024KarrasAutoGuidance}, we apply two commonly used metrics to assess the quality of the generated images. We calculate the Fréchet inception distance (FID) \cite{2017HeuselFID} and the Fréchet distance (FD) using DINOv2 \cite{2024OquabDINOv2} features, as done by Stein et al. \cite{2023SteinFD}. We follow Karras et al.'s terminology and call this last metric FD-DINOv2.

We generate 2000 images equally distributed between classes to use 10 images per class, using always the same set of random seeds. We then use a pretrained InceptionV3 \cite{2016SzegedyInceptionV3} and DINOv2 \cite{2024OquabDINOv2} to extract image features. We compare these features to those obtained with the same models applied to all images from Tiny ImageNet's training dataset.

\paragraph{Classification-based evaluation.}

To avoid any biases imposed by the use of a single family of metrics, we apply a pretrained classifier to the same 2000 generated images and calculate the Top-1 and Top-5 accuracy when predicting the ground-truth label out of the 200 classes available on Tiny ImageNet. 

Huynh et al. \cite{2022HuynhTinySwinClassifier} fine-tuned a Swin-L transformer on Tiny ImageNet, achieving a Top-1 accuracy of 91.35\% on Tiny ImageNet classification. The Swin-L model had previously been pre-trained on ImageNet-21k \cite{2021LiuSwin}. 

\section{Licences}

\begin{itemize}
    \item Autoguidance code \cite{2024KarrasAutoGuidance} \\ $\hookrightarrow$ Creative Commons BY-NC-SA 4.0 license
    \item EDM2 models \cite{2024KarrasEDM2} \\ $\hookrightarrow$ Creative Commons BY-NC-SA 4.0 license
    \item InceptionV3 model \cite{2016SzegedyInceptionV3} \\ $\hookrightarrow$ Apache 2.0 license
    \item DINOv2 model \cite{2024OquabDINOv2} \\ $\hookrightarrow$ Apache 2.0 license
    \item Tiny ImageNet Swin-L classifier \cite{2022HuynhTinySwinClassifier} \\ $\hookrightarrow$ Apache 2.0 license
    \item Tiny ImageNet dataset \cite{2015LeTinyImageNet, HuggingFaceTinyImageNet, KaggleTinyImageNet} \\ $\hookrightarrow$ Custom non-commercial license from ImageNet \cite{2009DengImageNet}
\end{itemize}

\section{Funding}

Funding for the research presented in this manuscript has been provided by University of Glasgow, Dotphoton A.G., and the Centre for Doctoral Training in Applied Photonics (CDTAP) funded by the Engineering and Physical Sciences Research Council (EPSRC) of UK Research and Innovation (UKRI).

\end{document}